%% file: main.tex
\documentclass[preprint,12pt,authoryear]{elsarticle}

\usepackage[left=4.2cm, right=4.2cm, marginparwidth=3cm, marginparsep=0.3cm]{geometry}
\usepackage[colorinlistoftodos,prependcaption,textsize=tiny, textwidth=3.0cm,disable]{todonotes}
\usepackage{marginnote}
\newcounter{todoside}
\newcommand{\altodo}[2][]{%
  \stepcounter{todoside}%
  \ifodd\value{todoside}%
    \normalmarginpar
  \else
    \reversemarginpar
  \fi
  \todo[#1]{#2}%
}

\usepackage{mathtools}
\usepackage{amsmath, amssymb, amsthm, bm}
\usepackage{hyperref}
\usepackage{float}
\usepackage{color}
\usepackage{xfrac}

\usepackage[ruled, vlined, linesnumbered]{algorithm2e}
\usepackage[T1]{fontenc}
\usepackage{graphicx}
\usepackage{subcaption}
\usepackage{booktabs}
\usepackage{multirow}
\usepackage{tabularx}
\usepackage{tikz}
\usetikzlibrary{arrows.meta, positioning, fit, backgrounds, calc}

\definecolor{cScen}{HTML}{534AB7}
\definecolor{cHaz}{HTML}{0F6E56}
\definecolor{cFld}{HTML}{993C1D}
\definecolor{cRule}{HTML}{9C9A92}
\definecolor{cText}{HTML}{5F5E5A}

\newtheorem{problem}{Problem}
\newtheorem{definition}{Definition}

\newtheorem{assumption}{Assumption}

\newcommand{\varSev}{S}                       % severity variable, S_h
\newcommand{\varOcc}{Z}                       % hazardous event occurrence, Z_h \in {0,1}
\newcommand{\varRisk}{\mathcal{R}}                  % risk description at a state, R(x)

\newcommand{\sethaz}{\mathcal{H}}             % set of hazardous events
\newcommand{\setsev}{\mathcal{S}}             % ordered set of severities
\newcommand{\setedge}{E}
\newcommand{\umeasure}{\mathcal{Q}}           % uncertainty measure
\newcommand{\knowledge}{\mathcal{K}}          % background knowledge
\newcommand{\setpath}{\Pi}
\newcommand{\haz}{h}                          % hazardous event label
\newcommand{\sev}{s}                          % severity value
\newcommand{\node}{v}
\newcommand{\nsev}{K}                         % number of severity states
\newcommand{\npath}{N}                        % number of paths in a path set

\newcommand{\opProb}{P}                         % probability operator
\newcommand{\prob}{p}                         % probability value
\newcommand{\evidence}{\mathcal{E}}           % evidence assignment, \evidence(\state)
\newcommand{\parentof}{\operatorname{par}}

\newcommand{\weight}{w}                       % consequence weight
\newcommand{\wmax}{\weight_{\max}}
\newcommand{\nweight}{\tilde{\weight}}        % normalized weight
\newcommand{\varrisk}{r}                      % risk index

\newcommand{\cutoff}{\varrisk_{\mathrm{cut}}}      % risk acceptance limit

\newcommand{\state}{x}
\newcommand{\stateset}{\mathcal{X}}
\newcommand{\workspace}{\mathcal{W}}
\newcommand{\dimw}{d}                         % workspace dimension
\newcommand{\opPosition}{\operatorname{pos}}
\newcommand{\position}{\zeta}
\newcommand{\defpath}{\sigma}
\newcommand{\edge}{e}
\newcommand{\edgelen}{\ell}
\newcommand{\arclen}{d}

\newcommand{\direction}{\theta}
\newcommand{\cost}{c}
\newcommand{\tree}{\mathcal{T}}
\newcommand{\nodeset}{V}
\newcommand{\costtogo}{g}
\SetKwFunction{ChooseParent}{ChooseParent}
\SetKwFunction{Rewire}{Rewire}
\SetKwFunction{Neighbor}{Neighbor}
\newcommand{\stepsize}{\eta}                  % steer step length
\newcommand{\goalrad}{\ell_{\mathrm{goal}}}   % goal acceptance radius
\newcommand{\neighborradius}{\ell_{\mathrm{nb}}}       % rewiring neighborhood radius
\newcommand{\nodebudget}{N_{\max}}              % node budget
\newcommand{\distweight}{\alpha}              % distance weight
\newcommand{\riskweight}{\beta}               % accumulated-risk weight
\newcommand{\turnweight}{\gamma}              % heading-change weight
\newcommand{\turnangle}{\phi}                 % turn angle at a node
\newcommand{\turnmax}{\turnangle_{\max}}      % hard turn-angle limit
\newcommand{\turncost}{\tau}                  % turn cost
\newcommand{\refleng}{L_{\mathrm{ref}}}       % turn reference length
\newcommand{\accrisk}{\rho}                  % accumulated risk index of an edge

\newcommand{\spacing}{\Delta_{\varrisk}}      % risk sample spacing
\newcommand{\nsamp}{n}                        % samples on one edge
\newcommand{\tinf}{t_{\varrisk}}              % mean wall-clock of one risk calculation
\newcommand{\nfeas}{\kappa}                   % fraction reaching compute_cost
\newcommand{\ncalls}{Q}                       % number of risk calculations (queries to the risk model)
\newcommand{\deadline}{T_{\text{max}}}                     % replanning deadline
\newcommand{\freearea}{A_{\text{free}}}                     % free area

\newcommand{\speed}{v_{\mathrm{d}}}           % desired speed
\newcommand{\lookahead}{d_{\mathrm{look}}}    % grounding lookahead distance
\newcommand{\shoremax}{d_{\mathrm{shore}}^{\max}}
\newcommand{\dcpa}{d_{\mathrm{CPA}}}
\newcommand{\dcpamax}{\dcpa^{\max}}
\newcommand{\tcpa}{t_{\mathrm{CPA}}}
\newcommand{\tcpamax}{\tcpa^{\max}}
\newcommand{\relvel}{v_{\mathrm{rel}}}
\newcommand{\relvelmax}{\relvel^{\max}}
\newcommand{\mindepth}{z_{\min}}              % depth hard limit
\newcommand{\depthmax}{z_{\max}}              % depth range
\newcommand{\opEdge}{\delta}
\newcommand{\opPeak}[2]{\max_{#2} #1}
\newcommand{\opInt}[2]{\int_{#2} #1}

\journal{Ocean Engineering}

\begin{document}

\begin{frontmatter}



    \title{Risk-Informed Path Planning: Collision and Grounding Avoidance Application for Autonomous Surface Vessels}


    \author[label1]{Emir Cem Gezer \corref{cor1}} %% Author name
    \ead{emir.cem.gezer@ntnu.no}

    \author[label2]{Ozan Kaya}
    % \ead{o.kaya@iat.rwth-aachen.de}

    \author[label1]{Ingrid B. Utne}
    \author[label1]{Roger Skjetne}
    \cortext[cor1]{Corresponding Author}

    \affiliation[label1]{organization={Department of Marine Technology (IMT), Norwegian University of Science and Technology (NTNU)},
        %% addressline={},
        city={Trondheim},
        postcode={NO-7052},
        %% state={},
        country={Norway}}
    %% Abstract
    \affiliation[label2]{organization={Chair of Information and Automation Systems for Process and Material Technology, RWTH Aachen University},
        %% addressline={},
        city={Aachen},
        postcode={D-52056},
        %% state={},
        country={Germany}}

    \begin{abstract}
        Maritime Autonomous Surface Ships (MASS) must demonstrate safety before and during operations, necessitating online risk assessment to guide decision-making at sea.
        This paper presents a risk-informed Rapidly Exploring Random Tree Star (RRT*) path planner that integrates a Bayesian Belief Network (BBN) as an online risk model.
        The BBN evaluates the probability and severity distribution of hazardous events, which are aggregated into a bounded scalar risk index.
        During the search, this risk index is evaluated at sample points along candidate edges.
        The planner optimizes a composite cost function comprising path distance, heading change penalties, and accumulated risk, subject to a hard constraint enforcing a risk acceptance limit.
        Each search tree vertex stores the BBN posterior at its state, enabling the tree to function as an explainable risk map for individual hazards.
        We define design requirements for the BBN to serve as the risk model in this context and introduce a runtime model that translates replanning deadlines into strict limits on single risk calculation times.
        The planner is demonstrated in a simulated harbor entry scenario involving collision and grounding hazards.
        Measured run times validate that the planner operates within the computational bounds predicted by the runtime model for a single replanning cycle.
    \end{abstract}

    \begin{graphicalabstract}
    \end{graphicalabstract}

    \begin{highlights}
        \item Risk-informed RRT* planner using probabilistic hazardous event mapping
        \item Multi-mode rewiring strategy for safety and path efficiency optimization
        \item Robust performance in dynamic and uncertain marine environments
        \item Enhanced collision and grounding avoidance capability
    \end{highlights}

    %% Keywords

\end{frontmatter}



\section{Introduction}
\label{sec:introduction}

The maritime domain is moving towards higher levels of autonomy as a means to make transportation safer, cleaner, and cheaper.
Autonomy is expected to reduce crew-related costs and to move part of the work to shore-based operations under remote monitoring \citep{kretschmann2017analyzing, alamoush2024drivers}.
Human error is reported to be a contributing factor to ship accidents.
Of the marine casualties and incidents investigated in the EU and EEA between 2015 and 2024, 54.2\% involved at least one human action as an accident event \citep{emsa2025annual}.
Consequently, removing crew from the bridge could mitigate some accidents \citep{wrobel2017towards}, but the accidents that still happen potentially lead to more severe consequences due to the lack of crew to mitigate and contain the damage.
One obstacle to MASS adaptation has been uncertainty regarding how autonomous or remotely operated vessels can be demonstrated to be safe \citep{alamoush2024drivers}.
To address the safety concerns, the International Maritime Organization (IMO) first conducted a scoping exercise \citep{IMO2021MASS}, then released a dedicated code for autonomous ships \citep{imo2026masscode}, and DNV has published guidelines to ensure the safety level is comparable to or better than that of crewed vessels \citep{DNV_CG_0264_2024}.
Both sets of documents require that a given operation is verified to be sufficiently safe.

Decision-making for a maritime vessel spans its entire life cycle, from design through operations to decommissioning.
Design determines the vessel's capabilities.
Operations unfold at multiple timescales: mission planning covers a single voyage, path planning and energy management run online over minutes, and feedback control loops correct for disturbances many times per second \citep{marley2020hybrid}.
Autonomy requires online risk analysis and modeling that runs at the timescale at which the vessel operates and is updated using vessel observations and operational information \citep{parhizkar2022online, rokseth2015dynamic, utne2020towards}.

A distinct family of methods contributes to handling safety at shorter timescales.
Representative examples are control barrier functions \citep{marley2021maneuvering}, model predictive control \citep{tengesdal2024obstacle}, and signal temporal logic \citep{krasowski2026pacstl}.
They encode constraints that keep the state inside a safe subset of the state space, satisfy a bounded temporal specification, or require an obstacle to be cleared within the planning horizon, corresponding respectively to CBFs, MPC, and STL.
These methods, however, do not consider the likelihood and the consequences of an event; namely, operational risk.
An online risk model built with a Bayesian belief network (BBN) can estimate risk, and the model can be evaluated at any point in the operating area.
This has been used to steer path planning towards safer routes \citep{bremnes2024holistic, kristensen_comparison_2024}, to evaluate risk across operational alternatives \citep{kristensen2025evaluating}, and to guide the search over risk scenarios \citep{maidana_dynamic_2024}.
A BBN can incorporate many risk-influencing factors (RIFs) and jointly capture both the likelihood and the consequences of hazardous events \citep{trucco2008bayesian, khakzad2011safety, hanninen2014bayesian}.

Several approaches have formalized risk in path planning.
\citet{lefebvre2016integration} integrated risk into a hierarchical planner for underwater vehicles, and \citet{chen2022risk} routed an underwater vehicle around an oil spill using a risk map.
\citet{zhen2023improved} modified A* for ships so that ocean current, water depth, and traffic-separation rules enter the cost.
\citet{fiskin2021fuzzy} combined a fuzzy ship domain with a search over candidate maneuvers to select collision-avoidance actions.
Closer to the present work, \citet{maidana2023risk} used BBN-derived hazardous-event probabilities to drive a path planner, and \citet{bremnes2024holistic} built a risk model for a marine robot that a planner can query at any point in the operating area.
\citet{kristensen_comparison_2024} compared several ways of providing the same risk information to a planner, and \citet{kristensen2025evaluating} showed that the choice of risk metric changes the resulting operational decisions.
Our earlier work \citep{kaya2026risk} used BBN-derived hazardous-event probabilities as a cost term in an RRT* planner, following the same BBN-to-planner pattern.
A recent survey of risk metrics for autonomous systems reports that the choice of risk metric remains open \citep{utne2025systematic}.

This paper presents a risk-informed path planner driven by an online risk model, designed so that the full probabilistic risk spectrum is available to the planner at its own timescale.
A Bayesian belief network (BBN) returns a severity distribution of consequences for each hazardous event, and a cost model condenses these distributions into a single bounded score called the \emph{risk index}.
The risk index is evaluated at sample points along each candidate edge; at each point, the BBN reads the relevant RIFs and produces the event-specific severity distributions that feed the cost model.
The maximum risk index along an edge is checked against a risk-acceptance limit, and an edge that exceeds the limit is rejected even if its mean risk index is low.
A heading-change penalty is added to the edge cost to minimize heading changes and sharp turns, leading to straighter paths between necessary turns.
We demonstrate that the presented risk-informed planner can operate in real-time, and we provide a study on how the tuning affects the computation time.
Beyond producing a path, the planner's search tree can be interpreted as a risk map for each hazardous event, which we use to explore the risk landscape of a harbor entry scenario.

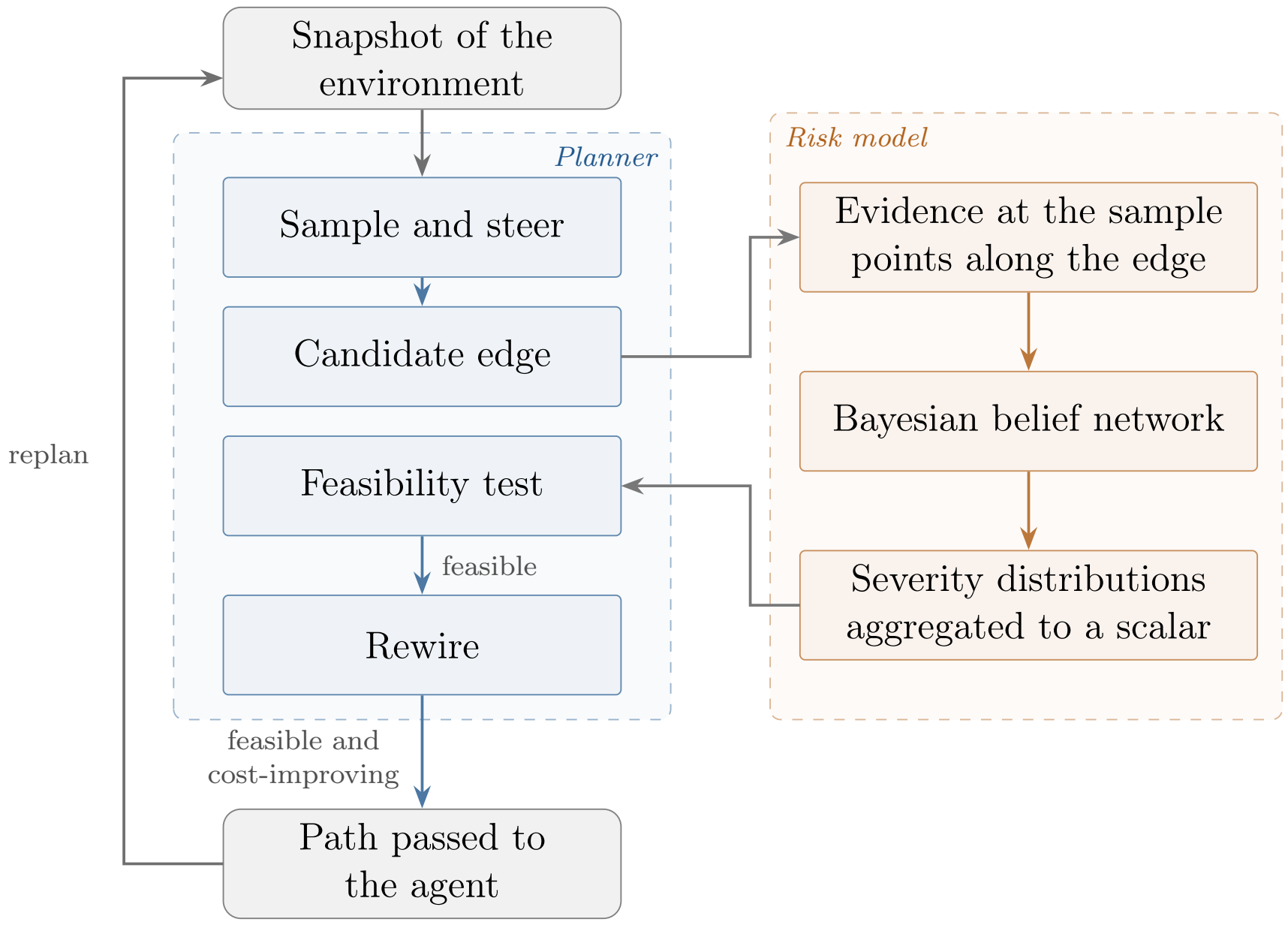
\begin{figure}[h]
    \centering
    \resizebox{0.9\linewidth}{!}{\input{diagrams/framework.tex}}
    \caption{ %
        The framework has two parts that run together in a loop.
        The risk model is queried at the states the planner considers, and it returns a risk index that the planner uses to compare candidate paths. %
    }
    \label{fig:framework}
\end{figure}

\altodo[color=green!40]{
    I removed the "overview of the framework" subsection. It looks much cleaner like this.
}
The proposed method builds on RRT* \citep{karaman2011sampling}, and each vertex of the tree stores the risk description obtained from a BBN (see Figure \ref{fig:framework}).
Storing probabilistic information at tree vertices is known from planning under uncertainty, where vertices hold a state distribution or a belief \citep{luders2010chance, bry2011rapidly}.
Here, the stored quantity is the probability of each hazardous event and the severity distribution of its consequences.
Risk-informed planners use risk either as a cost to minimize or as a constraint to satisfy \citep{majumdar2019should, blackmore2011chance, maidana2023risk}.
The proposed planner uses risk in both roles.
It stores the BBN posterior at every vertex, so the tree can be read afterwards as a risk map of the explored area.

The main contributions of this work are as follows:
\begin{itemize}
    \item
          A risk-informed RRT* planner where each vertex stores the risk description of its state.
          The search tree can therefore be read as a risk map for each hazardous event.

    \item
          A set of requirements that a BBN must meet to serve as the risk model, and a reduction of its output to a bounded risk index.
          The index grows with both the probability and the severity of a hazardous event.

    \item
          A cost evaluation that samples the risk index with the course angle of the edge and the time of arrival.
          The accumulated risk index enters the cost, and the peak value is held below a risk acceptance limit.

    \item
          A run-time model that turns a replanning deadline into a limit on the duration of one risk calculation.
          The model gives the largest node budget that a risk model can support, and it is checked in a simulated harbor entry.

\end{itemize}

The rest of the paper is organized as follows.
Section~\ref{sec:risk-modelling} explains how the probabilities and consequence severities of hazardous events are aggregated to a single risk index used as a planning cost.
Section~\ref{sec:planner} states the risk-informed, turn-penalized planning problem and explains the RRT*-style planner that solves it.
Section~\ref{sec:maritime-autonomy} explains how the risk model is applied to a maritime collision and grounding scenario.
Section~\ref{sec:navigation-study} presents a navigation study used to test the planner and discusses the results.
Section~\ref{sec:conclusions} concludes the paper.

\textbf{Mathematical notations}: The symbol $\lvert \cdot \rvert$ denotes the absolute value, the Euclidean norm, or the cardinality of a set, depending on its argument.
Calligraphic letters denote sets, upper-case letters random variables, and upright subscripts labels.
A superscript $(\cdot)^{*}$ marks an optimal quantity, a tilde indicates a normalized quantity, a bar means an average, and a hat gives the highest value along one edge, as in $\hat{\varrisk}_k$ for $\edge_k$ and $\hat{\varrisk}_\node$ for the incoming edge of $\node$.
Severity levels are ordered by $\prec$, and a quantity at a vertex $\node$ is evaluated at the state $\state_\node$ it holds.

\section{Risk Modeling}
\label{sec:risk-modelling}

Suppose the vessel is navigating within a state space $\stateset$, where each valid configuration is represented by a state $\state \in \stateset$.
The planner's objective is to find a sequence of configurations that forms a path.
To evaluate and optimize such paths, the planner requires a metric to compare the risk at any two states.
To achieve this, we design a risk model that provides the probability of each hazardous event, alongside a probability distribution over the severity of its consequences.
Because the search reaches states that were not known before it started, we introduce a scalar \emph{risk index} that is defined at every state the planner can query.
The index has two properties that make it usable as a cost.
First, it is bounded, so it has the same range at every state.
Second, it increases when a hazardous event becomes more likely or its consequences become more severe.

We follow the terminology of \citet{rausand2020riskassessment}.
A hazard is a source of danger that may cause harm to an asset, such as shallow water or another vessel.
A hazardous event is the first event in a sequence of possible events that, if not controlled, leads to harm, such as a grounding or a collision.
The consequences of a hazardous event are graded by their severity.
For each hazardous event $\haz$, the severity variable $\varSev_\haz$ is a random variable that grades the consequences that $\haz$ may have.
We call the distribution of $\varSev_\haz$ the severity distribution of $\haz$.
Severity variables take values in a common set of severity levels $\setsev \coloneqq \{\sev_1, \dots, \sev_\nsev\}$, ordered as $\sev_1 \prec \sev_2 \prec \cdots \prec \sev_\nsev$.
We compute a \emph{risk index} from the posterior severity distributions of the hazardous events in four steps.
First, a BBN returns a severity distribution for each hazardous event at the queried state.
Second, a consequence proxy assigns a number to each severity level.
Third, the expected value of the consequence proxy gives one value per hazardous event, which sets aside the spread of the distribution.
Fourth, a combination rule merges the hazardous events into one value per state, which treats them as independent.

\subsection{Risk and Risk Description}
\label{sec:method-definition}

We follow \citet{aven_risk_2009} and define risk as the uncertainty about the consequences of an activity, together with how severe those consequences are.
Risk in this sense belongs to the activity, and an analysis produces a \emph{risk description} for it.
A risk description is the four-tuple
\begin{align}
    \bigl( \sethaz,\; \setsev,\; \umeasure,\; \knowledge \bigr),
    \label{eq:risk-description}
\end{align}
where $\sethaz$ is the predefined set of hazardous events, $\setsev$ is the predefined set of severity levels, $\umeasure$ is a probability measure over pairs of hazardous event and severity level, and $\knowledge$ is the background knowledge that $\umeasure$ is based on.
The consequences enter the description only through their severity levels.
The analyst chooses the hazardous events and the severity levels, so the description covers only these.
The probabilities express a degree of belief given $\knowledge$, and not a counted frequency.
In this work, $\umeasure$ is computed with a BBN.
The background knowledge $\knowledge$ includes the structure of the network and the judgments behind its conditional probability tables.
\altodo{%
    I expanded the explanation of the background knowledge to include the structure of the network and the judgments behind its conditional probability tables.
}
The BBN does not express the uncertainty in this knowledge, and we discuss this as a limitation in Section~\ref{sec:conclusions}.

We evaluate the measure at a query state $\state$ and write the description there as
\begin{align}
    \varRisk(\state) \coloneqq \bigl( \sethaz,\; \setsev,\; \umeasure(\cdot \mid \state),\; \knowledge \bigr).
    \label{eq:risk-description-state}
\end{align}
The hazardous events and the severity levels are fixed, and only the measure changes with $\state$.
Evaluating $\umeasure(\cdot \mid \state)$ gives a probability for each pair of hazardous event and severity level.

\subsection{Bayesian Belief Network}

The probability of a hazardous event and the severity of its consequences depend on several RIFs that are only partly observable and that depend on each other.
We model these RIFs and their dependencies with a BBN.
Observable variables are given from what the system measures, and inference returns a posterior over the unobserved ones; see e.g. \citep{maidana2023risk}.

\begin{assumption}[Planner-compatible BBN]
    \label{ass:bbn}
    \altodo{I changed it from definition to assumption.}
    The BBN meets three requirements.
    First, a subset of its nodes, the evidence nodes, takes values that can be computed at any query state $\state \in \stateset$.
    Second, each hazardous event $\haz \in \sethaz$ has an occurrence node $\varOcc_\haz$ and is associated with at least one severity node $\varSev_\haz$ with states in $\setsev$.
    Third, $\varOcc_\haz$ is an ancestor of each severity node of $\haz$, and each severity node is a descendant of at least one evidence node.
\end{assumption}

\altodo{I removed the sentences explaining what the assumption \ref{ass:bbn} is for.}
Any BBN that meets Assumption~\ref{ass:bbn} can be used by the planner, and the network in Section~\ref{sec:maritime-autonomy} is one example.
Such a network should be designed to capture the risk spectrum from the agent's perspective for a given action under consideration.
We structure BBN with resemblance to the bow-tie model \citep{rausand2020riskassessment}.

\subsection{Posterior Severity Distribution}
\label{sec:method-bbn-output}

\altodo{i changed the title of the section}
Let $\evidence(\state)$ denote the assignment of values to the evidence nodes at $\state$, and let $\opProb(\cdot \mid \evidence(\state))$ denote the posterior probability measure that the BBN returns when queried with $\evidence(\state)$.
These probabilities give the measure $\umeasure(\cdot \mid \state)$ in \eqref{eq:risk-description-state}.
The planner uses only the posterior probabilities of the severity nodes,
\begin{align}
    \prob_{\haz,\sev}(\state) \coloneqq \opProb\bigl(\varSev_\haz = \sev \mid \evidence(\state)\bigr), \quad \haz \in \sethaz,\ \sev \in \setsev,
    \label{eq:scenario-posterior}
\end{align}
\altodo{
    This section is here to introduce the notation and to explain the meaning of the lowest severity level, not to present anything novel.
}%
which is the probability that the consequences of $\haz$ have severity level $\sev$, given the evidence at $\state$.
The lowest level $\sev_1$ stands for no or negligible consequences, and it includes the case where $\haz$ does not occur.
Severity nodes of this kind are also used in earlier BBNs for autonomous marine systems \citep{johansen2023development, bremnes2024holistic}.

\subsection{Consequence Proxy}
\label{sec:method-proxy}

We use a consequence proxy to compare severity levels on a numerical scale.
\altodo[color=green!40]{
    Severity distribution $\varSev$ takes values from Severity levels $\sev$ and is represented with a severity node in the BBN.
}
The severity levels are ordered labels, and they do not express intervals of a loss.
To recover a numerical scale, we assign a number to each level with a \emph{consequence proxy} $\weight_\haz : \setsev \to \mathbb{R}_{\geq 0}$,
\begin{align}
    0 = \weight_\haz(\sev_1) \le \weight_\haz(\sev_2) \le \cdots \le \weight_\haz(\sev_\nsev),
    \label{eq:consequence-weights}
\end{align}
where $\weight_\haz(\sev)$ is the weight of severity level $\sev$.
We call it a proxy because it simplifies the risk calculation.
We take the severity distribution that the BBN gives and map it to numbers, instead of computing a higher-fidelity risk metric such as the value at risk, CVaR, or EVaR \citep{rockafellar2000optimization,ahmadi2012entropic}.
\altodo[color=green!40]{
    I've added my reasoning behind why I chose the `consequence proxy'. It wasn't there before.
}
The values $\weight_\haz(\sev)$ are design parameters.
The lowest level $\sev_1$ stands for no or negligible consequences, so we set $\weight_\haz(\sev_1) = 0$.

\subsection{Event risk index}

To create a single comparable value for the planner, we reduce the severity distribution of each hazardous event to one number.
\altodo{The output of the BBN is multiplied by a weight. After this, I call it the risk index.}
Let $\nweight_\haz(\sev) \coloneqq \weight_\haz(\sev)/\wmax$ denote the normalized weight, where $\wmax$ is the largest weight over all hazardous events.
We define the \emph{event risk index} $\varrisk_\haz : \stateset \to [0,1]$ of hazardous event $\haz$ as the expected normalized weight of $\varSev_\haz$ given the evidence at $\state$, that is,
\begin{align}
    \varrisk_\haz(\state) \coloneqq \mathbb{E}\bigl[\nweight_\haz(\varSev_\haz) \mid \evidence(\state)\bigr] = \sum_{\sev \in \setsev} \prob_{\haz,\sev}(\state)\, \nweight_\haz(\sev).
    \label{eq:hazard-risk-index}
\end{align}

A hazardous event, $\haz$, may harm several assets, such as people, the environment, and the vessel.
The network can then hold one severity variable for each asset, $\varSev_\haz^{\text{asset}}$, each with its own consequence proxy.
The event risk index is then the sum of the expected weights over these variables, divided by the sum of their largest weights.
It stays in $[0,1]$, and with a single severity variable it aggregates to \eqref{eq:hazard-risk-index}.

\subsection{Aggregated Risk Index}
\label{sec:method-risk-cost-combine}
\altodo{I changed the title of the section from "Combining hazardous events" to "Aggregated Risk Index".}

We combine the event risk indices of all hazardous events into one risk index $\varrisk : \stateset \to [0, 1]$, that is,
\begin{align}
    \varrisk(\state) \coloneqq 1 - \prod_{\haz \in \sethaz}\bigl(1 - \varrisk_\haz(\state)\bigr).
    \label{eq:cost-model-combined-risk}
\end{align}
The risk index is zero only when all event risk indices are zero, and it is one when any event risk index is one.
Every hazardous event adds to the risk index, and it increases with each $\varrisk_\haz(\state)$.
The event risk indices are expected severity weights, so $\varrisk(\state)$ is a score.
Note that the rule treats the hazardous events as independent, although the BBN links them through common parents.

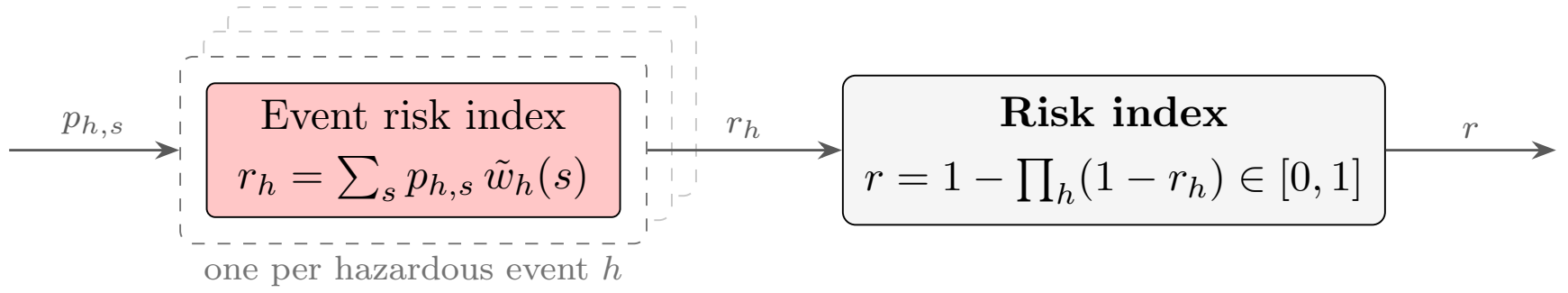
\begin{figure}[H]
    \centering
    \resizebox{0.9\linewidth}{!}{\input{diagrams/generic_severity_structure.tex}}
    \caption{Aggregating the posterior severity probabilities into the risk index.}
    \label{fig:generic_severity_structure}
\end{figure}

\section{Risk-Informed Path Planning}
\label{sec:planner}

\subsection{Problem definition}

Let $\stateset$ be the state space and let $\state \in \stateset$ denote a state.
A state holds the position of the vessel together with any other information the risk model might need, such as its heading, velocity, travel duration, and mode of operation.
\altodo{It just says $d=2$ now. Much cleaner.}
The planner searches for a path in a workspace $\workspace \subset \mathbb{R}^{\dimw}$ of positions, where we restrict our development to $\dimw = 2$ in this work.
The free space $\workspace_{\mathrm{free}} \subseteq \workspace$ is the set of obstacle-free positions.
The map $\opPosition \colon \stateset \to \workspace$ gives the position of each state $\state$.

\begin{definition}[Edge]
    \label{def:edge}
    For $\state_{k-1}, \state_{k} \in \stateset$ with positions $\position_k \coloneqq \opPosition(\state_k)$,
    the edge $\edge_k \coloneqq (\position_{k-1}, \position_{k})$ is associated with the straight-line segment $\opEdge_k$ parametrized by $\lambda$, that is,
    \begin{align}
        \opEdge_k \colon [0,1] \to \workspace,
        \qquad
        \opEdge_k(\lambda) \coloneqq (1-\lambda)\, \position_{k-1} + \lambda\, \position_{k}.
        \label{eq:edge-param}
    \end{align}
    The edge $\edge_k$ is admissible if and only if $\opEdge_k(\lambda) \in \workspace_{\mathrm{free}}$ for all $\lambda \in [0,1]$.
\end{definition}

For an edge $\edge_k$ and $\lambda \in [0,1]$, let $\state_k(\lambda) \in \stateset$ denote the state at the position $\opEdge_k(\lambda)$.
Its heading is the direction of $\edge_k$, and its time is the arrival time at $\opEdge_k(\lambda)$.

\begin{definition}[Edge evaluation]
    \label{def:edge-eval}
    For any bounded $f \colon \stateset \to \mathbb{R}$ that is piecewise constant along each edge $\edge_k$, define the maximum and the line integral of $f$ along $\opEdge_k$ as
    \begin{align}
        \opPeak{f}{\edge_k} \coloneqq \max_{\lambda \in [0,1]} f\big(\state_k(\lambda)\big),
        \quad
        \opInt{f}{\edge_k} \coloneqq \edgelen_k \int_{0}^{1} f\big(\state_k(\lambda)\big) \, d\lambda ,
        \label{eq:edge-eval}
    \end{align}
    where $\edgelen_k \coloneqq \lvert \position_{k} - \position_{k-1} \rvert$ is the Euclidean length of $\edge_k$.
    \altodo{$\position \in \mathbb{R}^2, d=2$, so, there is no complications left there anymore.}
\end{definition}

A path is a finite ordered sequence of admissible edges
\begin{align}
    \defpath \coloneqq (\edge_1, \edge_2, \dots, \edge_{\npath}).
\end{align}
Each edge carries three quantities: the length $\edgelen_k$, the accumulated risk index $\accrisk_k$, and the heading-change penalty $\turncost_k$, which combine into the edge cost and the additive path cost
\begin{align}
    c_k \coloneqq \distweight \edgelen_k + (1-\distweight) \riskweight \accrisk_k + \turnweight \turncost_k,
    \quad
    C(\defpath) \coloneqq \sum_{k=1}^{\npath} c_k,
    \quad
    \distweight \in [0,1], \, \riskweight, \turnweight \geq 0 .
    \label{eq:planner-edge-cost}
\end{align}
An edge is risk-admissible if and only if
\begin{align}
    \hat{\varrisk}_k \coloneqq \opPeak{\varrisk}{\edge_k} \leq \cutoff,
    \qquad \cutoff \in [0,1],
    \label{eq:risk-gate}
\end{align}
where $\cutoff$ is the \emph{risk acceptance limit}, that is, the highest risk index the planner accepts at any point of a path.
This test on $\hat{\varrisk}_k$ ensures that a short stretch with a high risk index cannot be averaged away by an otherwise low-risk edge.
Note that the risk acceptance limit is a design parameter of the planner.

\begin{problem}{Risk-informed, turn-penalized path planning.}
\label{prob:risk-informed-planning}
Given $\workspace_{\mathrm{free}}$, a risk index field $\varrisk$, weights $\distweight, \riskweight, \turnweight$,
a risk acceptance limit $\cutoff$, and positions $\position_{\mathrm{init}}, \position_{\mathrm{goal}} \in \workspace_{\mathrm{free}}$, find
\begin{align}
     & \defpath^{*} = \underset{\defpath \in \setpath}{\arg\min} \; C(\defpath),
    \\
     & \setpath \coloneqq \{ \defpath :
    \position_0 = \position_{\mathrm{init}}, \, \position_{\npath} = \position_{\mathrm{goal}},\, \hat{\varrisk}_k \leq \cutoff, k = 1, \dots, \npath
    \}.
    \label{eq:planner-problem}
\end{align}
\end{problem}

\subsection{Accumulated Risk Index of an Edge}

Scoring an edge with a risk index from a single point leaves the rest of the edge invisible to the cost.
\altodo{We have risk acceptance criteria on the peak risk index, but not on the accumulated risk. Accumulated risk enters as an optimization variable.}
The accumulated risk index has the unit of length,
\begin{align}
    \accrisk_k = \opInt{\varrisk}{\edge_k}.
    \label{eq:edge-risk-exposure}
\end{align}

In practice, $\accrisk_k$ and $\hat{\varrisk}_k$ are evaluated at $n_k$ sample points spaced evenly along the edge,
\begin{align}
    \lambda_i = \frac{i}{n_k}, \quad i = 1, \dots, n_k,
    \qquad
    n_k = \min\Bigl( \max\bigl( \lceil \edgelen_k / \spacing \rceil, 1 \bigr), n_{\max} \Bigr),
    \label{eq:samples}
\end{align}
\altodo{%
    The accumulated risk is meant to grow with path length. Edge length itself does not matter. Risk index enters as a cost term, so the planner tries to reduce it as well.
}
where $\spacing$ is the target spacing and $n_{\max}$ caps the number of samples on one edge.
The start point $\lambda = 0$ is left out, since it is the end point of the previous edge and has already been evaluated.
The maximum and the integral in \eqref{eq:edge-eval} are then replaced by the maximum and a Riemann sum over the samples,
\begin{align}
    \hat{\varrisk}_k \approx \max_{i} \varrisk\bigl(\state_k(\lambda_i)\bigr),
    \qquad
    \accrisk_k \approx \frac{\edgelen_k}{n_k} \sum_{i=1}^{n_k} \varrisk\bigl(\state_k(\lambda_i)\bigr).
    \label{eq:edge-discrete}
\end{align}
Setting $n_k = 1$ scores the edge at its end point.

\subsection{Heading-Change Penalty}

Minimizing distance and the accumulated risk index alone can produce paths that zigzag through low-risk regions.
We therefore add a heading-change penalty that grows as the turn gets sharper.

Let $\direction_k \coloneqq \position_k - \position_{k-1}$.
The absolute turn angle between $\edge_{k-1}$ and $\edge_k$ is
\begin{align}
    \turnangle_k = \bigl| \operatorname{atan2}\bigl( \direction_{k-1} \times \direction_k, \direction_{k-1} \cdot \direction_k \bigr) \bigr| \in [0,\pi],
    \label{eq:turn-angle}
\end{align}
where $\direction_{k-1} \times \direction_k$ is the 2D scalar cross product of two planar vectors.
The first edge of a path has no incoming edge, so $\turnangle_1 = 0$.
The turn cost is
\begin{align}
    \turncost_k =
    \begin{cases}
        \infty,                                & \turnangle_k > \turnmax, \\
        \dfrac{\turnangle_k}{\pi} \, \refleng, & \text{otherwise},
    \end{cases}
    \label{eq:turn-cost}
\end{align}
where $\turnmax \in [0,\pi]$ is an optional hard limit on the turn angle.
The reference length $\refleng$ states the penalty in the same unit as edge length, so a full $180^\circ$ turn costs as much as $\refleng$ of straight path.
The weight $\turnweight$ of \eqref{eq:planner-edge-cost} trades off against distance and the accumulated risk index in the same way.
\altodo{$turnweight$ is introduced at \eqref{eq:planner-edge-cost} and explained here.}
\subsection{Planner}

Problem~\ref{prob:risk-informed-planning} is solved by an RRT*-style search (Algorithm~\ref{alg:rrt-star}).
The search builds a tree $\tree = (\nodeset, \setedge)$, where each vertex $\node \in \nodeset$ holds a state $\state_\node \in \stateset$.
We write $\opPosition(\node)$ for $\opPosition(\state_\node)$.
The root $\node_0$ holds the initial state, with $\opPosition(\node_0) = \position_{\mathrm{init}}$.
Every other vertex has a single parent $\parentof(\node) \in \nodeset$ and is reached through the incoming edge
\begin{align}
    \edge_\node = \bigl( \opPosition(\parentof(\node)),\, \opPosition(\node) \bigr).
    \label{eq:incoming-edge}
\end{align}
Each vertex stores the record
\begin{align}
    \bigl( \parentof(\node),\; \costtogo(\node),\; \arclen(\node),\; \hat{\varrisk}_\node,\; \opProb(\cdot \mid \evidence(\state_\node)) \bigr),
    \label{eq:vertex}
\end{align}
where $\parentof(\node)$ is the parent, $\costtogo(\node)$ is the accumulated cost from the root, $\arclen(\node)$ is the accumulated path length from the root, and $\hat{\varrisk}_\node \coloneqq \opPeak{\varrisk}{\edge_\node}$ is the highest risk index along the incoming edge.
The last entry is the full posterior that the BBN returns at $\state_\node$, over all of its nodes.
The path quantities follow from those of the parent,
\begin{subequations}
    \label{eq:vertex-recursion}
    \begin{align}
        \costtogo(\node) & = \costtogo(\parentof(\node)) + c_\node,      & \text{Cost to come},     \\
        \arclen(\node)   & = \arclen(\parentof(\node)) + \edgelen_\node, & \text{Distance to come},
    \end{align}
\end{subequations}
where $c_\node$ and $\edgelen_\node$ are the cost and length of $\edge_\node$, and $\costtogo(\node_0) = \arclen(\node_0) = \hat{\varrisk}_{\node_0} = 0$.

The connectable set $\nodeset_{\mathrm{c}} \subseteq \nodeset$ holds the root and every vertex that has a parent.
The function $\Neighbor(\position)$ returns the vertices in $\nodeset_{\mathrm{c}}$ that lie within the radius $\neighborradius$ of the position $\position$.
The search minimizes $\costtogo$ when it selects a parent or rewires a neighbor, while it keeps $\hat{\varrisk}_\node$ at or below the risk acceptance limit $\cutoff$.

For a sample point on an edge leaving $\node$, the path length $\arclen(\node)$ plus the distance along the edge gives the distance the agent has traveled.
Divided by the transit speed, this distance gives the time at which the agent is predicted to reach the sample point.
The dynamic obstacles are placed at their predicted positions at that time before the risk index is evaluated.
The posterior is taken at the last sample point of $\edge_\node$, which is $\opPosition(\node)$.
It is saved in the vertex, so that the planner can later ask which hazardous event drove the risk index at $\node$.
The tree is therefore a sampled risk map of the region the search is exploring.
The planner can also return a path together with its risk description.

\begin{algorithm}[h]
    \footnotesize
    \KwIn{$\position_{\mathrm{init}}$, $\position_{\mathrm{goal}}$, $\workspace_{\mathrm{free}}$, $\varrisk(\cdot)$, $\alpha,\beta,\gamma$, $\cutoff$}
    \KwOut{$\tree$, and $\defpath^\ast$ if $\parentof(\node_{\mathrm{goal}}) \ne \varnothing$}
    Create $\node_0$ at $\position_{\mathrm{init}}$ with $\parentof(\node_0) \leftarrow \varnothing$ and $\costtogo(\node_0) = \arclen(\node_0) = \hat{\varrisk}_{\node_0} = 0$\;
    Create $\node_{\mathrm{goal}}$ at $\position_{\mathrm{goal}}$ with $\parentof(\node_{\mathrm{goal}}) \leftarrow \varnothing$ and $\costtogo(\node_{\mathrm{goal}}) \leftarrow \infty$\;
    $\nodeset \leftarrow \{\node_0\}$, \ $\mathrm{reach} \leftarrow \textbf{false}$\;
    \Repeat{$|\nodeset| \ge \nodebudget \wedge \mathrm{reach}$}{
        $\position_{\mathrm{rand}} \sim \mathcal{U}(\workspace_{\mathrm{free}})$, \ $\node_{\mathrm{near}} \leftarrow \operatorname{nearest}(\nodeset_{\mathrm{c}}, \position_{\mathrm{rand}})$\;
        $\position_{\mathrm{new}} \leftarrow \operatorname{steer}(\opPosition(\node_{\mathrm{near}}), \position_{\mathrm{rand}})$\;
        \lIf{$(\opPosition(\node_{\mathrm{near}}), \position_{\mathrm{new}})$ is not admissible}{\textbf{continue}}
        $\node_{\mathrm{par}} \leftarrow \ChooseParent(\position_{\mathrm{new}})$\;
        Add $\node_{\mathrm{new}}$ at $\position_{\mathrm{new}}$ to $\nodeset$\;
        \lIf{$\node_{\mathrm{par}} = \varnothing$}{$\parentof(\node_{\mathrm{new}}) \leftarrow \varnothing$, \textbf{continue}}
        $\parentof(\node_{\mathrm{new}}) \leftarrow \node_{\mathrm{par}}$, and set the record of $\node_{\mathrm{new}}$ with \eqref{eq:vertex-recursion}\;
        $\Rewire(\node_{\mathrm{new}}, \Neighbor(\opPosition(\node_{\mathrm{new}})))$\;
        \If{$\lvert\opPosition(\node_{\mathrm{new}}) - \position_{\mathrm{goal}}\rvert \le \goalrad$ \textbf{and} $(\opPosition(\node_{\mathrm{new}}), \position_{\mathrm{goal}})$ is admissible}{
            $\mathrm{reach} \leftarrow \textbf{true}$, \ $\Rewire(\node_{\mathrm{new}}, \{\node_{\mathrm{goal}}\})$\;
        }
    }
    \caption{Risk-informed RRT*}
    \label{alg:rrt-star}
\end{algorithm}
\begin{algorithm}[h]
    \footnotesize
    \SetKwProg{Fn}{Function}{:}{}
    \Fn{\ChooseParent{$\position$}}{
        $\node^\ast \leftarrow \varnothing$, \ $\costtogo^\ast \leftarrow \infty$\;
        \ForEach{$\node \in \Neighbor(\position)$}{
            $\edge \leftarrow (\opPosition(\node), \position)$\;
            \lIf{$\edge$ is not admissible $\vee\ \opPeak{\varrisk}{\edge} > \cutoff$}{\textbf{continue}}
            \lIf{$\costtogo(\node) + c(\edge) < \costtogo^\ast$}{$\node^\ast \leftarrow \node$, \ $\costtogo^\ast \leftarrow \costtogo(\node) + c(\edge)$}
        }
        \KwRet $\node^\ast$\;
    }
    \caption{Parent selection}
    \label{alg:parent_selection}
\end{algorithm}

\begin{algorithm}[h]
    \footnotesize
    \SetKwProg{Fn}{Function}{:}{}
    \Fn{\Rewire{$\node$, $\nodeset_{\mathrm{nb}}$}}{
        \ForEach{$\node_{\mathrm{n}} \in \nodeset_{\mathrm{nb}} \setminus \{\node, \parentof(\node)\}$}{
            $\edge \leftarrow (\opPosition(\node), \opPosition(\node_{\mathrm{n}}))$\;
            \lIf{$\edge$ is not admissible $\vee\ \opPeak{\varrisk}{\edge} > \cutoff$}{\textbf{continue}}
            \If{$\costtogo(\node) + c(\edge) < \costtogo(\node_{\mathrm{n}})$}{
                $\parentof(\node_{\mathrm{n}}) \leftarrow \node$\;
                Set the record of $\node_{\mathrm{n}}$ with \eqref{eq:vertex-recursion}\;
            }
        }
    }
    \caption{Rewiring}
    \label{alg:rewiring}
\end{algorithm}

\subsubsection{Hard feasibility by orphaning}
The planner might sample a node that does not satisfy the risk acceptance limit $\hat{\varrisk}_k \le \cutoff$.
In this case, the newly created node is removed from the tree by \emph{orphaning}, but the node is kept in $\nodeset$ to retain information.
To \emph{orphan} a node, we simply set its parent to $\varnothing$.
The same test also prevents orphan nodes from being reconnected during rewiring.
As a result, an orphaned node is excluded from the connectable set and can never be included in a returned path.

\subsubsection{Cost-driven connection and rewiring}
Both \ChooseParent and \Rewire use the cost-to-come $g(\node) + c(\edge)$ of \eqref{eq:planner-edge-cost} as cost (See Algorithms \ref{alg:rewiring} and \ref{alg:parent_selection}).
\ChooseParent selects the best parent for $\node_{\mathrm{new}}$ to the vertex in $\nodeset$ that minimizes this quantity, and \Rewire reassigns an existing neighbor to the new parent whenever routing through it lowers that neighbor's cost-to-come.

\subsection{Real-time requirements}

Almost all of the planner's run time is spent on BBN inference.
Therefore, the run time mainly depends on two things: how long one risk calculation takes, and how many risk calculations the planner makes.
We write the run time as
\begin{equation}
    T_{\mathrm{run}} \approx \ncalls_{\mathrm{tot}} \, \tinf ,
    \label{eq:runtime}
\end{equation}
where $\tinf$ is the mean duration of one risk calculation and $\ncalls_{\mathrm{tot}}$ is the count of risk calculations in one run.
The duration $\tinf$ covers building the evidence, inference in the BBN, and computing the risk index.
It depends on the size and structure of the network and on the hardware.
The count $\ncalls_{\mathrm{tot}}$ depends on the \emph{node budget}, $\nodebudget$ and the number of neighbors within $\neighborradius$.
These two factors can be measured separately, and we treat them one at a time below.

We first count the risk calculations.
The planner grows a tree until it holds $\nodebudget$ vertices.
Each new vertex is connected to the vertices already in the tree that lie within the radius $\neighborradius$.
With uniform sampling, the average number of such neighbors in a tree of size $\lvert\nodeset\rvert$ is \citep{karaman2011sampling}
\begin{align}
    \bar{m}\bigl(\lvert\nodeset\rvert\bigr) = \frac{\pi \neighborradius^{2}}{\freearea} \lvert\nodeset\rvert ,
    \label{eq:neighbors}
\end{align}
where $\freearea$ is the free area.
Each neighbor gives two candidate edges, one as a possible parent of the new vertex and one for rewiring through it.
A share $\nfeas \in (0,1]$ of these edges reaches the risk model, as the rest are blocked by obstacles or already known to exceed the risk acceptance limit $\cutoff$.
Each remaining edge is sampled at $\bar{\nsamp}$ points on average, where $\bar{\nsamp}$ is the average number of samples per edge, and each sample is one risk calculation.
Summing over the tree sizes $\lvert\nodeset\rvert = 1, \dots, \nodebudget$ gives
\begin{equation}
    \ncalls_{\mathrm{tot}} \approx \sum_{\lvert\nodeset\rvert=1}^{\nodebudget} 2\, \nfeas\, \bar{\nsamp}\, \bar{m}\bigl(\lvert\nodeset\rvert\bigr) \approx \nfeas\,\bar{\nsamp}\,\frac{\pi \neighborradius^{2}}{\freearea}\,\nodebudget^{2} .
    \label{eq:total}
\end{equation}
The count grows with the square of the node budget, because each new vertex finds more neighbors as the tree branches are populated.
The sampling-based planning literature offers several ways to lower the count \citep{karaman2011sampling,solovey2020critical}.

Inference cost cannot be estimated based on the structure of a BBN in any general way, because both exact and approximate inference in a BBN are NP-hard \citep{cooper1990computational,dagum1993approximating}.
Instead, we give a target on the duration itself, which a reader can measure directly for their own network.
The deadline $\deadline$ is the time the planner has to produce a new plan before the current one stops being usable.
It is bounded by the shortest time constant the agent should be able to act on.
All risk calculations of one run must fit inside $\deadline$, which gives a requirement on the risk model,
\begin{equation}
    \tinf \le \frac{\deadline}{\ncalls_{\mathrm{tot}}} .
    \label{eq:budget}
\end{equation}
With $\ncalls_{\mathrm{tot}}$ from Equation~\eqref{eq:total}, the largest node budget that a given risk model can support is
\begin{equation}
    \nodebudget \le \sqrt{\frac{\deadline \, \freearea}{\nfeas\,\bar{\nsamp}\,\pi\,\neighborradius^{2}\,\tinf}} .
    \label{eq:max-budget}
\end{equation}

\subsection{Risk Map from the Search Tree}
\label{sec:risk-map}
\altodo{I added a new subsection briefly describing how the risk map is generated from the search tree.}
Each vertex stores the BBN inference results to create a risk map of the explored region.
The vertex values, including the orphaned vertices, are interpolated onto a regular grid and smoothed with a Gaussian filter.
The resulting map shows which hazardous event and which RIFs drove the risk, which explains why the planner chose one route over another.
In addition, it can be passed to other systems, such as a remote control center or a supervisory risk controller \citep{utne2020towards, johansen_supervisory_2022}.
The value at a vertex depends on the heading and arrival time of its incoming edge, so the map shows the risk along the explored vertices and not a property of the position alone.

\section{Risk-Informed Path Planning for Maritime Autonomy}
\label{sec:maritime-autonomy}
We demonstrate the proposed planner framework on a maritime navigation task.
A vessel transits a confined waterway with shallow water and other vessels.
These hazards can lead to grounding and collision, and the planner must produce a path that is acceptable with respect to both.

\subsection{Bayesian Belief Network}

\begin{figure}[h]
    \centering

    \altodo{I choose this BBN figure for multiple reasons. I can use the BBN with libraries other than genie; the priors are not adjusted properly, so that bar chart with priors can be misleading, and I like this figure aesthetically more than the genie one.}

    \begin{minipage}[c]{\textwidth}
        \begin{subfigure}[c]{\linewidth}
            \centering
            % \resizebox{!}{0.31\textheight}{\input{diagrams/bbn_without_evidence.tex}}
            \resizebox{0.9\linewidth}{!}{\input{diagrams/bbn_without_evidence.tex}}
            % \caption{}
            \label{fig:application_bbn_without_evidence}
        \end{subfigure}
    \end{minipage}\hfill \\[1.2ex]
    \begin{minipage}[c]{\textwidth}
        \centering
        \begin{subfigure}[c]{0.38\linewidth}
            \centering
            \resizebox{!}{0.104\textheight}{\input{diagrams/bbn_encounter_properties.tex}}
            % \resizebox{\linewidth}{!}{\input{diagrams/bbn_encounter_properties.tex}}
            \label{fig:application_bbn_sub_encounter}
        \end{subfigure}%\\[1.2ex]
        \begin{subfigure}[c]{0.31\linewidth}
            \centering
            \resizebox{!}{0.104\textheight}{\input{diagrams/bbn_environment_properties.tex}}
            % \resizebox{\linewidth}{!}{\input{diagrams/bbn_environment_properties.tex}}
            % \caption{}
            \label{fig:application_bbn_sub_environment}
        \end{subfigure}%\\[1.2ex]
        \begin{subfigure}[c]{0.31\linewidth}
            \centering
            \resizebox{!}{0.104\textheight}{\input{diagrams/bbn_waterway_properties.tex}}
            % \resizebox{\linewidth}{!}{\input{diagrams/bbn_waterway_properties.tex}}
            % \caption{}
            \label{fig:application_bbn_sub_waterway}
        \end{subfigure}
    \end{minipage}
    \altodo[color=green!40]{I added weights from RankedNodes on the edges. This is rather similar to sensitivity analysis that one could get from GeNIe.}
    \caption{
        The BBN used in this work.
        The main structure is shown at the top, and the dashed nodes (a)--(c) are expanded below it into their evidence subtrees for encounter, environment, and waterway properties.
        The value on each connection is the weight of that parent in the ranked node it points to.
    }
    \label{fig:application_bbn}
\end{figure}
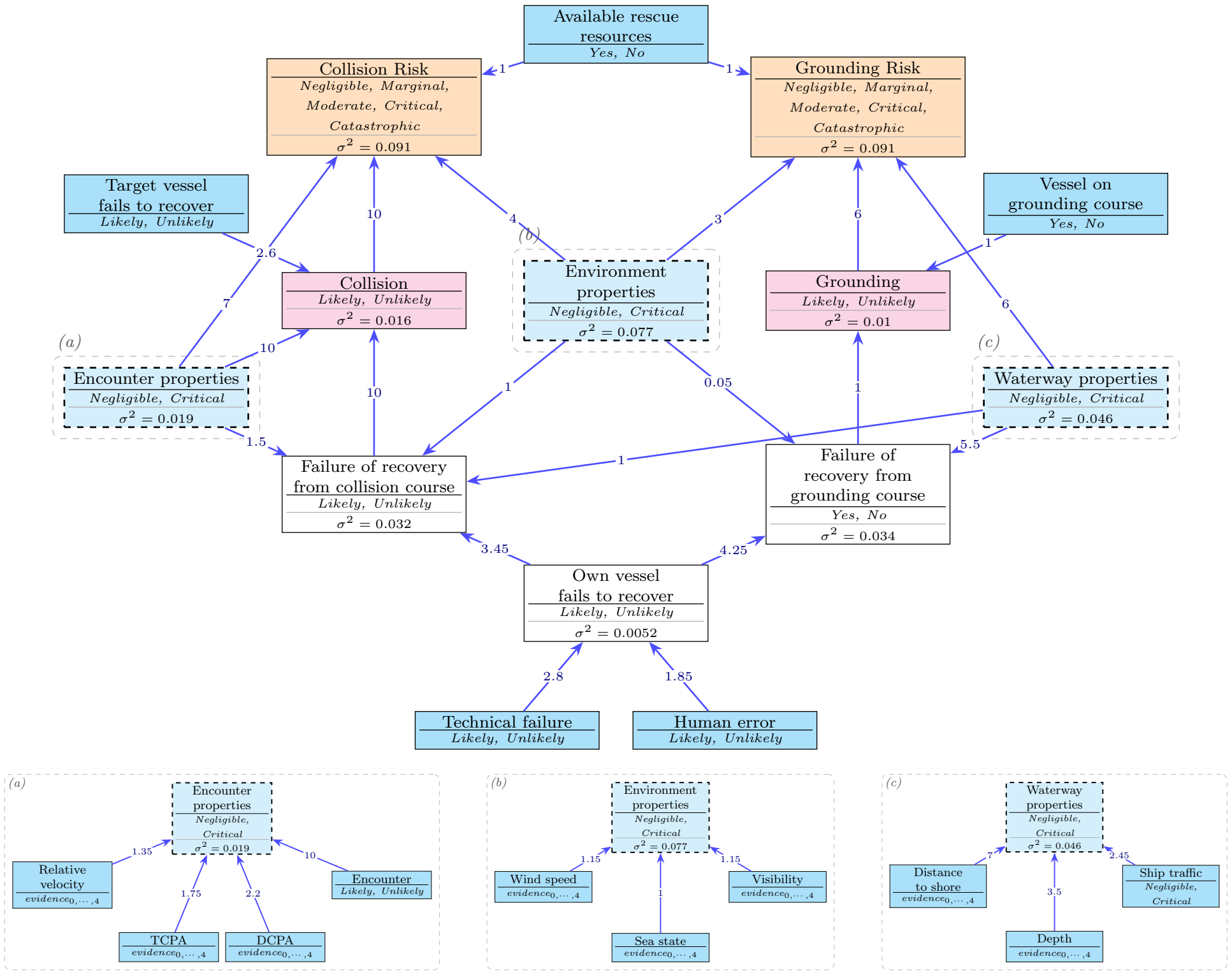

The BBN in Figure~\ref{fig:application_bbn} answers a question from the planner's point of view: \textit{If the own vessel passes this point on the heading of the edge and at the predicted time, how likely is a hazardous event, and how severe would its consequences be?}
The evidence describes the action under consideration, not the vessel's present state, so the network is queried at every sample point of every candidate edge.

\paragraph{Network Structure}
The BBN is a further development of the networks of \citet{maidana2023risk} and \citet{kristensen_comparison_2024}, adapted to path planning and extended with a severity node for each hazardous event similar to \citep{bremnes2024holistic}.
At the base, a set of observable RIFs describes the vessel's immediate situation: the geometry of any developing encounter, the waterway being transited, and the prevailing environmental conditions.
Aggregation nodes condense each subtree of evidence into a single assessment of whether that aspect of the situation is \emph{Negligible} or \emph{Critical}.
These feed the recovery nodes, which act as preventive barriers and express whether the vessel can still avoid a hazardous event by means other than the planner itself.
Their output determines the hazardous event nodes, and each hazardous event node is in turn the parent of a severity node describing how severe its consequences would be.
The available rescue resources act as a barrier on the consequence side, as they reduce the severity of the consequences of a collision or grounding.
The BBN used in this work can be found here \citep{emir_cem_gezer_2026_24502046}.
\altodo{CPTs will be provided in the supplementary material.}

\paragraph{Forming the conditional probability table}
We define the conditional probability tables of all non-observable nodes using ranked nodes \citep{fenton_using_2007}, which generate a whole table from a function of the parents, one weight per parent, and a variance.
Ranked nodes let us express expert knowledge in a simple way.
Here, each weight indicates how strongly a parent pushes the node up or down and the variance expresses how much uncertainty there is in that influence.
They can be built from expert knowledge alone, being easy to tune given the low number of parameters.
The weights and variances of the ranked nodes in this work are given in Figure~\ref{fig:application_bbn}.

\paragraph{Populating the evidence during search}
The evidence quantities are divided into position-dependent properties, such as waterway and environmental conditions, and encounter-dependent properties, such as \emph{distance at} and \emph{time to} closest point of approach (DCPA and TCPA), relative velocity, and the headings of the own vessel and dynamic targets.
DCPA and TCPA are used to characterize encounter geometry, while relative velocity describes encounter dynamics. The course of each edge defines the heading of the own vessel and is used to assess whether the vessel is on a grounding course.
Encounter geometry and dynamics are calculated based on the predicted courses and speeds of dynamic targets, the desired speed of the own vessel, and the accumulated path length.
In this work, we assume that the own vessel's speed is constant and that the dynamic targets maintain their current course and speed over each planning horizon.

\paragraph{State ordering convention}
Evidence state labels are ordered by the magnitude of the quantity they represent.
For example, a lower TCPA value corresponds to a lower-indexed label, and a higher relative velocity corresponds to a higher-indexed label.

\paragraph{Discretization of the evidence}
Each continuous numerical value is mapped onto an evidence label with uniform binning.
Uniform bins are used for their simplicity.
Values beyond the upper range are assigned to the corresponding highest state, so that each range defines where the information is significant.
The ranges are chosen as the distance at which the corresponding hazard becomes irrelevant for the decision.
Encounter evidence is computed based on the predicted course and the speed of the dynamic targets, the desired speed of the own vessel, and the accumulated path length.
When multiple encounters are possible, the encounter with the smallest DCPA is chosen as the most critical.

\subsection{Parameters}

Table~\ref{tab:parameters} lists the parameters of the planner, grouped by what they control.
The step length $\stepsize$, goal radius $\goalrad$, neighborhood radius $\neighborradius$, and node budget $\nodebudget$ set how the tree grows.
A larger $\nodebudget$ or $\neighborradius$ gives a smoother path and a denser risk map, but increases the number of risk calculations as shown in \eqref{eq:total}.
The weights $\distweight$ and $\riskweight$ set the trade-off between distance and the accumulated risk index in \eqref{eq:planner-edge-cost}, and the risk acceptance limit $\cutoff$ sets the highest risk index the planner accepts at any point.
The turn weight $\turnweight$, turn limit $\turnmax$, and reference length $\refleng$ set how strongly the planner avoids sharp turns.
The sample spacing $\spacing$ and the cap $n_{\max}$ set how finely risk is sampled along an edge, which trades accuracy of $\hat{\varrisk}_k$ and $\accrisk_k$ against run time.
The remaining parameters belong to the risk model.

\begin{table}[H]
    \centering
    \scriptsize
    \caption{Parameters of the planner and the risk model, with the values used in the case study.
        The same consequence proxy is used for collision and grounding.}
    \label{tab:parameters}
    \renewcommand{\arraystretch}{1.15}
    \begin{tabular}{@{} c @{\hspace{2.0em}} c @{}}
        \toprule
        % ---------------- Left half ----------------
        \begin{tabular}[t]{@{} l l r@{\,}l @{}}
            Symbol            & Meaning                 & \multicolumn{2}{l@{}}{Value}       \\
            \midrule
            \multicolumn{4}{@{}l}{\textit{Tree growth}}                                      \\
            \cmidrule(r){1-4}
            $\stepsize$       & Steer step length       & $50$                         & m   \\
            $\goalrad$        & Goal radius             & $75$                         & m   \\
            $\neighborradius$ & Neighborhood radius     & $150$                        & m   \\
            $\nodebudget$     & Node budget             & $1500$                       &     \\
            \addlinespace[2pt]
            \multicolumn{4}{@{}l}{\textit{Cost and acceptance}}                              \\
            \cmidrule(r){1-4}
            $\distweight$     & Distance weight         & $0.1$                        &     \\
            $\riskweight$     & Accumulated-risk weight & $100$                        &     \\
            $\cutoff$         & Risk acceptance limit   & $0.15$                       &     \\
            \addlinespace[2pt]
            \multicolumn{4}{@{}l}{\textit{Heading-change penalty}}                           \\
            \cmidrule(r){1-4}
            $\turnweight$     & Turn weight             & $0.6$                        &     \\
            $\turnmax$        & Turn-angle cutoff       & $\pi/4$                      & rad \\
            $\refleng$        & Turn reference length   & $50$                         & m   \\
            \addlinespace[2pt]
            \multicolumn{4}{@{}l}{\textit{Risk sampling}}                                    \\
            \cmidrule(r){1-4}
            $\spacing$        & Risk sample spacing     & $25$                         & m   \\
            $n_{\max}$        & Samples per edge, cap   & $3$                          &
        \end{tabular}
         &
        % ---------------- Right half ----------------
        \begin{tabular}[t]{@{} l l r@{\,}l @{}}
            Symbol       & Meaning                 & \multicolumn{2}{l@{}}{Value}       \\
            \midrule
            \multicolumn{4}{@{}l}{\textit{Vessel and evidence}}                         \\
            \cmidrule(l){1-4}
            $\speed$     & Desired speed           & $3$                          & m/s \\
            $\mindepth$  & Depth hard limit        & $6$                          & m   \\
            $\lookahead$ & Grounding lookahead     & $150$                        & m   \\
            $\shoremax$  & Distance-to-shore range & $150$                        & m   \\
            $\depthmax$  & Depth range             & $200$                        & m   \\
            $\dcpamax$   & DCPA range              & $100$                        & m   \\
            $\tcpamax$   & TCPA range              & $120$                        & s   \\
            $\relvelmax$ & Relative-velocity range & $5$                          & m/s \\
            \addlinespace[2pt]
            \multicolumn{4}{@{}l}{\textit{Consequence proxy} $\weight_\haz(\sev)$}      \\
            \cmidrule(l){1-4}
            $\sev_1$     & Negligible              & $0$                          &     \\
            $\sev_2$     & Marginal                & $4$                          &     \\
            $\sev_3$     & Moderate                & $16$                         &     \\
            $\sev_4$     & Critical                & $64$                         &     \\
            $\sev_5$     & Catastrophic            & $256$                        &
        \end{tabular}
        \\
        \bottomrule
    \end{tabular}
    \stepcounter{todoside}
    \altodo{%
        The values are set by judgment, with each level four times worse than the level below.
        They are not calibrated to accident data.
        I added this to the limitations.
    }
\end{table}

\section{Case study: Navigating into the Harbor}
\label{sec:navigation-study}

\begin{figure*}[h]
    \centering
    \begin{subfigure}[t]{0.50\textwidth}
        \centering
        \includegraphics[width=\textwidth]{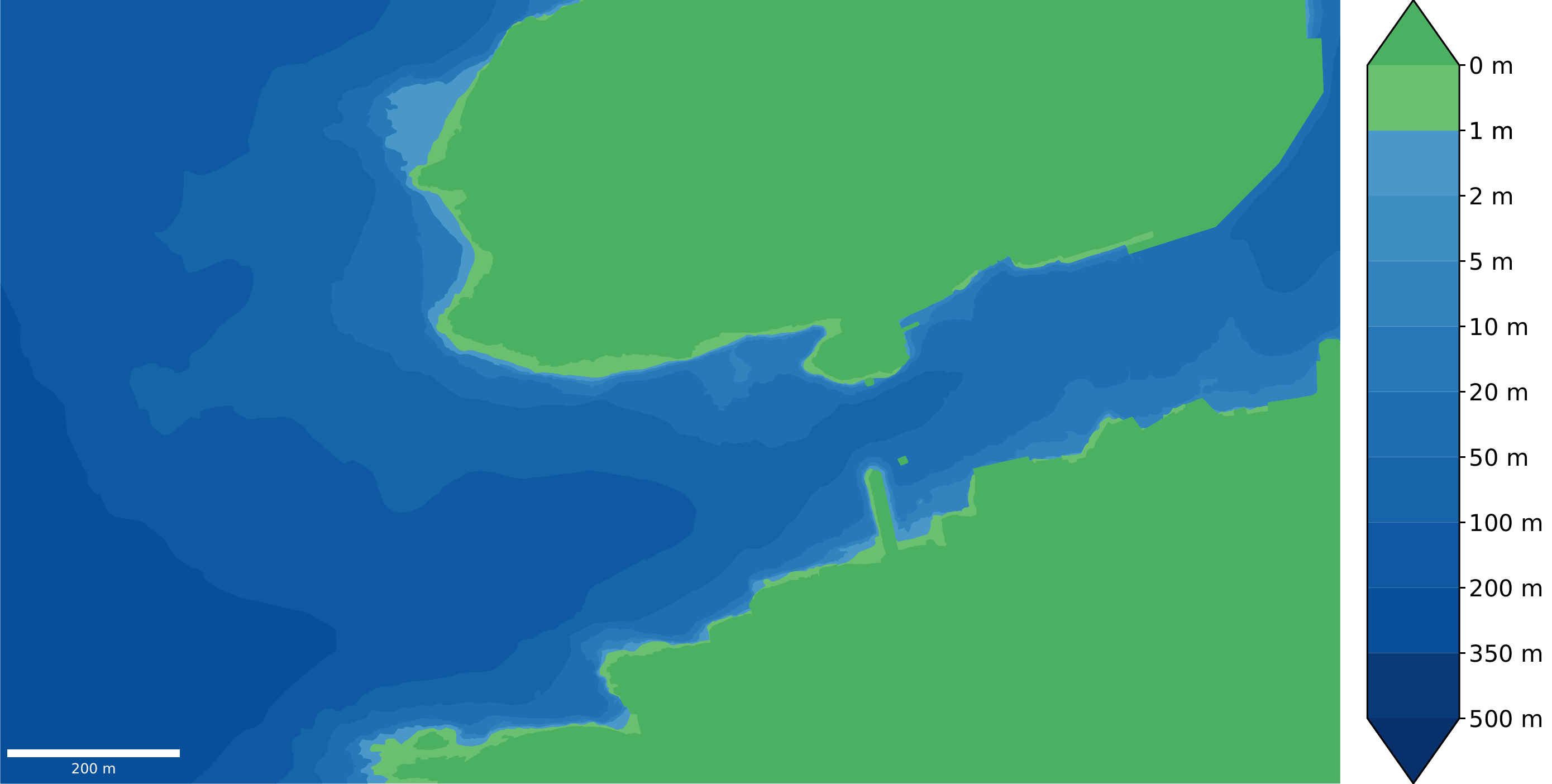}
        \caption{Kristiansund port west entry.}
        \label{fig:kristiansund_port_west_entry}
    \end{subfigure}%
    ~
    \begin{subfigure}[t]{0.40\textwidth}
        \centering
        \includegraphics[width=\textwidth]{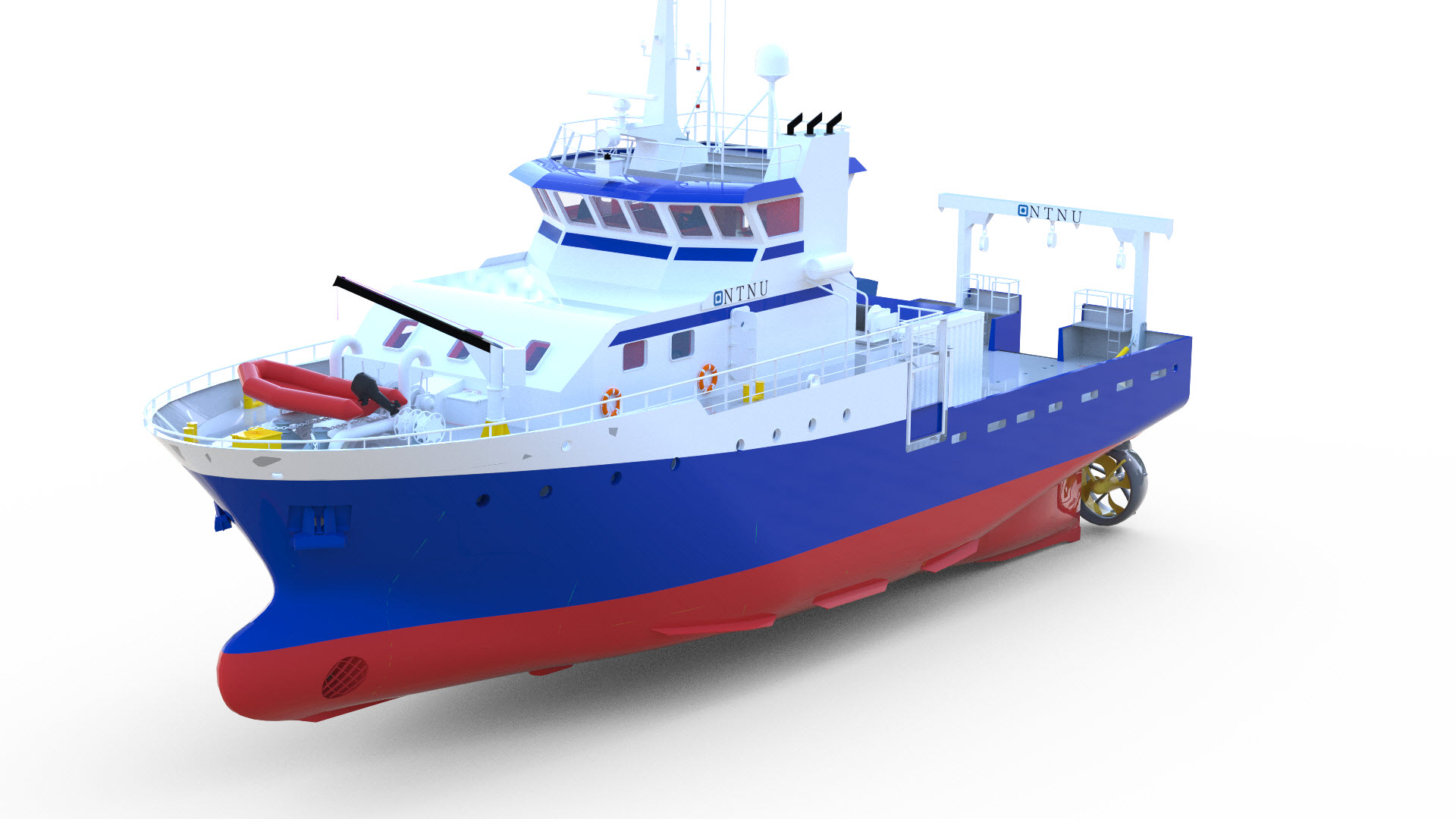}
        \caption{R/V Gunnerus.}
        \label{fig:gunnerus_illustration}
    \end{subfigure}
    \caption{%
        Test area and vessel used in this work.
        (\subref{fig:kristiansund_port_west_entry}) Bathymetric map of the west entry to Kristiansund, a narrow passage with shallow water on both sides.
        (\subref{fig:gunnerus_illustration}) Illustration of the research vessel R/V Gunnerus, which serves as the own ship in the scenarios. Courtesy of \citet{kleppe2026intelligent}.%
    }
    \label{fig:area_and_vessel}
\end{figure*}

We conducted a simulation study of a vessel entering the harbor of Kristiansund, Norway.
The own ship is modeled on NTNU's research vessel R/V Gunnerus, which is $36.25$\,m long and $9.9$\,m wide.
We use its dimensions and assume similar maneuvering capabilities, but the study does not include the real vessel or its dynamics.
Like Gunnerus, the simulated vessel is assumed to be fully actuated at low speed, so the planner may assume that stopping and turning on the spot are feasible.
The vessel transits at a constant speed $\speed = 3$\,m/s, which converts the path length into a time of arrival and places the dynamic targets when the vessel reaches each point.
Water shallower than $\mindepth = 6$\,m is treated as non-navigable and excluded from sampling.
All parameter values are listed in Table~\ref{tab:parameters}, and the discretization of the evidence is shown in Table~\ref{tab:discretization}.
We kept the same consequence proxy for both collision and grounding for simplicity, but the framework allows for different proxies for different hazards.

\begin{table}[H]
    \centering
    \caption{Discretization of the observable quantities onto BBN states. Values of the ranges are given in Table~\ref{tab:parameters}.}
    \label{tab:discretization}
    { \scriptsize
        \begin{tabular}{llll}
            \toprule
            Node              & Quantity                     & Range             & \# States \\
            \midrule
            DCPA              & relevant DCPA                & $[0, \dcpamax]$   & 5         \\
            TCPA              & relevant TCPA                & $[0, \tcpamax]$   & 5         \\
            Relative velocity & $\lvert \Delta v \rvert$     & $[0, \relvelmax]$ & 5         \\
            Distance to shore & distance to nearest obstacle & $[0, \shoremax]$  & 5         \\
            Depth             & relevant depth               & $[0, \depthmax]$  & 5         \\
            Ship traffic      & traffic density              & $[0, 1]$          & 2         \\
            Grounding course  & lookahead free-path test     & $\lookahead$      & 2         \\
            \bottomrule
        \end{tabular}
    }
\end{table}

% \begin{table}[H]
%     \centering
%     \scriptsize
%     \caption{Values used in the reported experiments. Symbols are defined in Table~\ref{tab:parameters}.}
%     \label{tab:experiment_parameters}
%     \begin{tabular}{l r@{\,}l}
%         \toprule
%         \multicolumn{3}{l}{\textit{Planner}}                \\
%         \addlinespace[2pt]
%         $\stepsize$       & $50$    & m                     \\
%         $\goalrad$        & $75$    & m                     \\
%         $\neighborradius$ & $150$   & m                     \\
%         $\nodebudget$     & $1500$  &                       \\
%         $\distweight$     & $0.1$   &                       \\
%         $\riskweight$     & $100$   &                       \\
%         $\cutoff$         & $0.05$  &                       \\
%         \midrule
%         \multicolumn{3}{l}{\textit{Heading-change penalty}} \\
%         \addlinespace[2pt]
%         $\turnweight$     & $0.6$   &                       \\
%         $\turnmax$        & $\pi/4$ & rad                   \\
%         $\refleng$        & $50$    & m                     \\
%         \midrule
%         \multicolumn{3}{l}{\textit{Risk sampling}}          \\
%         \addlinespace[2pt]
%         $\spacing$        & $25$    & m                     \\
%         \bottomrule
%     \end{tabular}
% \end{table}

\paragraph{Interpretation of the consequence proxy.}
In the case study, the severity level of an event is its worst outcome for people, the environment, or the vessel.
The lowest level, negligible, covers events with no collision or grounding, or light contact without damage.
Marginal events cause only cosmetic damage, and the vessel can continue its voyage.
Moderate events damage the hull, propeller, or rudder and cause at most minor injuries.
Critical events include a hull breach, loss of propulsion, serious injuries, or a small fuel spill, and the vessel needs a tug.
The highest level, catastrophic, covers loss of life, total loss of the vessel, or a large oil spill in the harbor.
With the weights in Table~\ref{tab:parameters}, the planner treats one catastrophic outcome as equal to four critical ones.

\begin{figure*}[h]
    \centering
    \begin{subfigure}[t]{0.62\textwidth}
        \vspace{0pt}
        \centering
        \includegraphics[width=\textwidth]{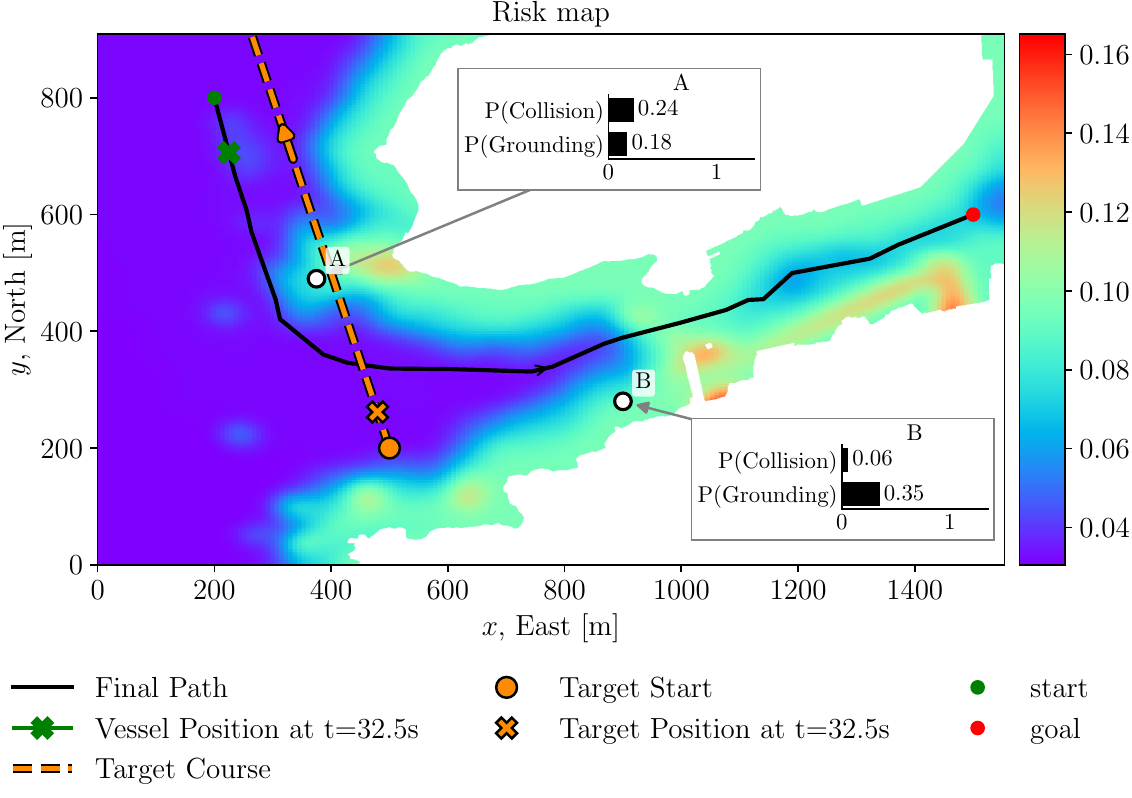}
        \caption{}
        \label{fig:search_explainability_risk}
    \end{subfigure}%
    \hfill
    \begin{minipage}[t]{0.35\textwidth}
        \vspace{0pt}
        \begin{subfigure}{\textwidth}
            \centering
            \includegraphics[width=\textwidth]{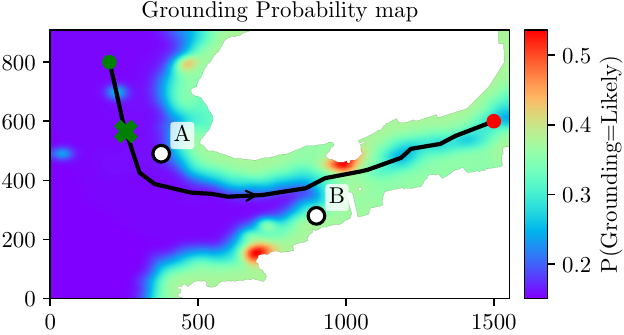}
            \caption{}
            \label{fig:search_explainability_grounding}
        \end{subfigure}
        \\
        \begin{subfigure}{\textwidth}
            \centering
            \includegraphics[width=\textwidth]{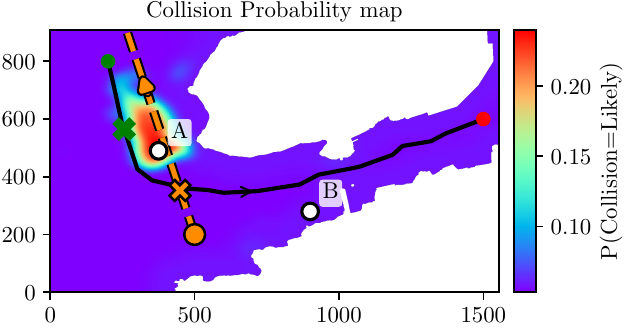}
            \caption{}
            \label{fig:search_explainability_collision}
        \end{subfigure}
    \end{minipage}
    \caption{%
        The risk map can be inspected to see which of the hazardous events are most likely to occur in a given location.
        (\subref{fig:search_explainability_risk}) Combined risk map used by the planner. The bar charts show the posterior probabilities of the hazardous event nodes at the selected points.
        (\subref{fig:search_explainability_grounding}) Grounding posterior as map.
        (\subref{fig:search_explainability_collision}) Collision posterior as map.%
    }
    \label{fig:search_explainability}
\end{figure*}

\paragraph{Explainability of the results}
The values in Figure~\ref{fig:search_explainability} are posterior probabilities of the state \emph{Likely} of the hazardous event nodes.
\altodo{%
    The snapshot and the sensitivity analysis answer different questions. Should I state this more explicitly?
}
They show the posterior probability of a hazardous event at a given location, conditional on the vessel taking the move under consideration.
The network was not calibrated to accident data, so the values compare states with each other.
Hence, they are not accident frequencies for the area.
\altodo{%
    The figure is a snapshot: the posterior of the "Likely" state at one state and time, given the move under evaluation, with no exposure period. The network is not calibrated to accident data, so the values are only relative and are not comparable to grounding statistics for the area. I added this to the figure text and the limitations.
}

\paragraph{Continuous planning}

\begin{figure*}[h]
    \centering
    \includegraphics[width=0.9\textwidth]{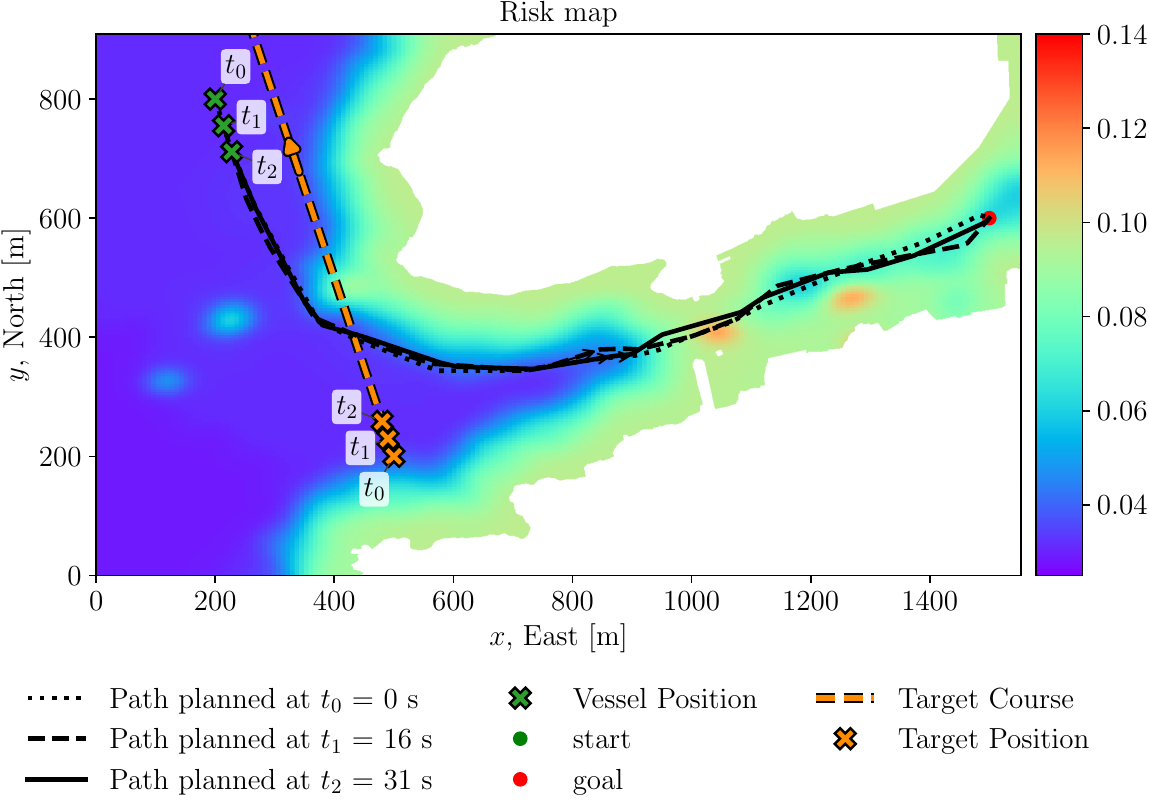}
    \caption{%
        Three successive plans from the same transit.
        A new search is started once the previous plan has been passed to the vessel, and each search uses a fresh snapshot of the surroundings.
        The route and the risk map both shift as the other vessel moves, so the plan stays matched to the current situation. $\nodebudget = 1200$ for all three searches.%
    }
    \label{fig:continuous_planning}
\end{figure*}

To handle a changing environment, the vessel continuously updates its plan as new information arrives.
The vessel follows the current plan while the next one is computed, and the next plan must be ready within $\deadline$  before the current one stops being usable.
For this reason, a new search starts as soon as the previous one ends.
Each search starts from the most recent snapshot of the surroundings.
Figure~\ref{fig:continuous_planning} shows three plans from the same transit, taken one after another.
The route changes as the other vessel moves through the area, and the risk map is updated accordingly.

\begin{figure}[h]
    \centering
    \includegraphics[width=0.75\linewidth]{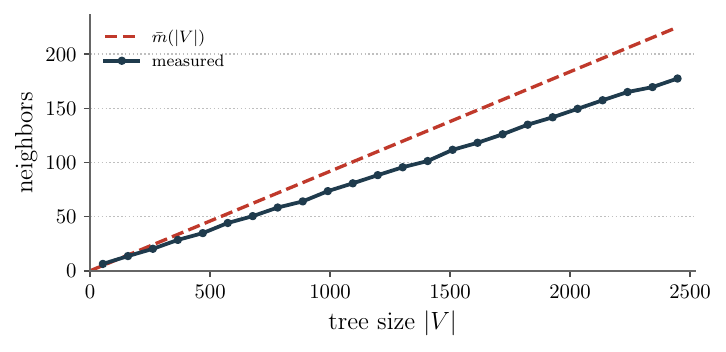}
    \caption{%
        Number of neighbors as the tree grows, over three runs, against $\bar{m}(\lvert\nodeset\rvert)$ of Equation~\eqref{eq:neighbors}.
        Shoreline clipping keeps the measured count below the prediction.%
    }
    \label{fig:crowding-neighbors}
\end{figure}

\begin{figure}[h]
    \centering
    \includegraphics[width=0.75\linewidth]{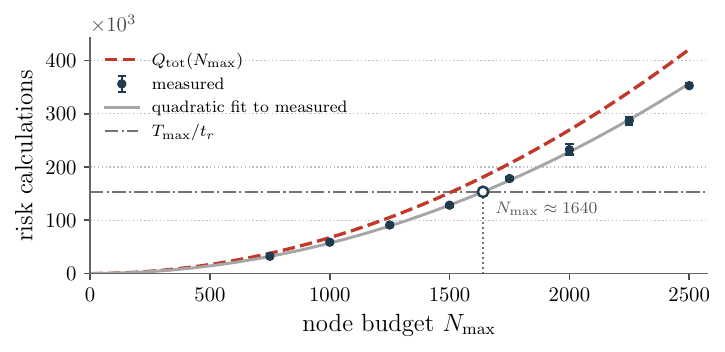}
    \caption{%
        Total number of risk calculations per run against the node budget, with $\ncalls_{\mathrm{tot}}(\nodebudget)$ from Equation~\eqref{eq:total} and a least-squares fit.
        The horizontal line is $\deadline/\tinf$, and the fit crosses it at $\nodebudget \approx 1640$; budgets to the left of the crossing fit one replanning cycle.%
    }
    \label{fig:crowding-risk-calculations}
\end{figure}

\paragraph{Computation time}
Figures~\ref{fig:crowding-neighbors} and~\ref{fig:crowding-risk-calculations} show that the two factors behind Equation~\eqref{eq:total} behave as predicted.
The number of neighbors grows linearly in the tree size, and so does the number of risk calculations spent on each new node, which makes the count over a whole run quadratic in the node budget.
The measured neighborhood stays about a fifth below the prediction, because discs centered near the shore are clipped by land.
Equation~\eqref{eq:total} is therefore an upper bound, which is the useful direction for sizing a system.

A single risk calculation takes $\tinf \approx 0.19$~ms, and its mean stays between 0.191 and 0.202~ms for node budgets from 750 to 2500, since one calculation does not depend on how many nodes are already in the tree.
Rearranged, Equation~\eqref{eq:budget} requires $\ncalls_{\mathrm{tot}} \le \deadline/\tinf$, which is the horizontal line in Figure~\ref{fig:crowding-risk-calculations}.
For the deadline $\deadline = \stepsize/\speed \approx 16.7$~s that follows from Table~\ref{tab:parameters}, the measured count crosses it at $\nodebudget \approx 1640$.
The risk calculations of one plan fit inside a replanning cycle below that budget.

\section{Conclusions}
\label{sec:conclusions}

This paper presented a risk-informed RRT* planner for autonomous surface vessels, driven by a Bayesian belief network that is evaluated online during the search.
We stated the requirements a network must meet to serve as the risk model, and reduced its severity distributions to a bounded risk index.
The risk index is sampled along each candidate edge with the heading of the edge and the time of arrival, so the same position can carry a different risk index depending on how and when the vessel reaches it.
The highest risk index on an edge is held below a risk acceptance limit, and a heading-change penalty avoids abrupt changes in direction.
Each vertex stores the BBN posterior behind its risk index, so the search tree also serves as a risk map that can be inspected for each hazardous event.
A simple run-time model turns a replanning deadline into a limit on the duration of one risk calculation, and it predicted the largest node budget that fit a replanning cycle in a simulated harbor entry at Kristiansund.

% Some limitations are worth mentioning.
The ranked node weights and variances were set by the authors.
A more systemic approach to eliciting expert judgements, e.g., by using the Delphi method \citep{dalkey_experimental_1963}, could potentially lead to adjustments of these weights and variance, and should be subject to further work.
Building a detailed and calibrated network is outside the scope of this paper, and any network that meets Assumption~\ref{ass:bbn} can replace this one without changes to the planner.
Moreover, the planner does not consider the International Regulations for Preventing Collisions at Sea (COLREG); hence, it may plan to pass a vessel on the side that the rules forbid if the risk there is low.
In addition to that, a path computed with a finite node budget is feasible but not necessarily optimal \citep{karaman2011sampling}, and the sampled peak along an edge can underestimate the true peak.
Finally, the case study was run in simulation, with straight path segments and target vessels that keep their course and speed.

% Several parts of the framework can be improved.
The network can be refined and calibrated from data or from a structured expert study, and it can replace the current one without changes to the planner \citep{hanninen2014bayesian}.
Severity nodes for people and the environment can be added to the network, each with its own consequence proxy.
Other risk measures can be explored in place of the expected consequence proxy, such as the value at risk, the conditional value at risk, or the entropic value at risk \citep{ahmadi2012entropic,rockafellar2000optimization}.
Each gives a different reading of the same severity distributions, for example by giving more weight to severe but unlikely outcomes.
The COLREG can be brought into the network itself, since a maneuver that other vessels do not expect adds risk even when the DCPA is large.
Object-oriented BBNs can keep a larger network manageable, as they split it into smaller parts that can be built and checked separately \citep{weidl2005applications}.
The planner can also be made faster, for example with a neighborhood radius that shrinks as the tree grows \citep{karaman2011sampling,solovey2020critical}.

\section*{Data availability}
The source code, data, and the parameters needed to reproduce the results are openly available on Zenodo at \url{https://doi.org/10.5281/zenodo.24502045}.

\section*{Acknowledgement}

Gezer and Skjetne acknowledge funding from the Research Council of Norway (RCN) through SFI AutoShip (RCN grant 309230), the Norwegian Maritime AI Centre (RCN grant 359242), and NTNU AMOS (RCN grant 223254).
Kaya and Utne's contributions are funded by the European Union.
Views and opinions expressed are, however, those of the author(s) only and do not necessarily reflect those of the European Union or the European Research Council Executive Agency.
Neither the European Union nor the granting authority can be held responsible for them.
This work is supported by ERC grant (BREACH, ID: 101142277, DOI 10.3030/101142277.)

\includegraphics[width=0.4\linewidth]{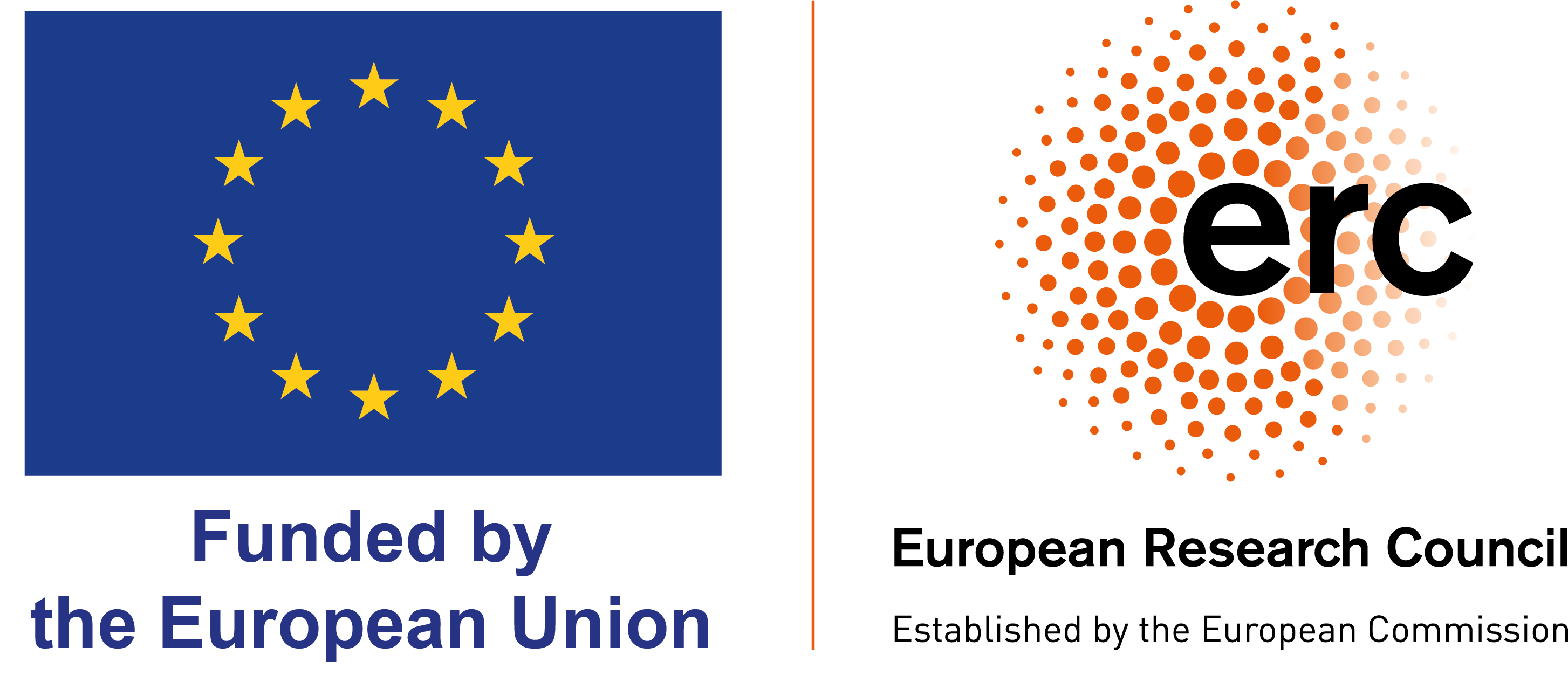}

\newpage

\appendix

\bibliographystyle{elsarticle-harv}
\bibliography{references}

\end{document}

%% file: diagrams/framework.tex
% ---------------------------------------------------------------------------
% Framework overview: the planner, the risk model, and the loop they run in.
% Self-contained: needs only \usepackage{tikz} in the preamble.
% Intended use: \resizebox{0.9\textwidth}{!}{\input{diagrams/framework.tex}}
% ---------------------------------------------------------------------------
\usetikzlibrary{arrows.meta}

\definecolor{fwplanner}{RGB}{31,86,140}
\definecolor{fwrisk}{RGB}{176,96,24}
\definecolor{fwagent}{RGB}{75,75,75}

\begin{tikzpicture}[
    font=\small,
    >={Stealth[length=2.3mm]},
    base/.style={
            draw, rounded corners=1.5pt, align=center,
            minimum height=10mm, inner xsep=3mm, inner ysep=1.4mm
        },
    pbox/.style={base, text width=34mm, draw=fwplanner!70, fill=fwplanner!7},
    rbox/.style={base, text width=40mm, draw=fwrisk!70,    fill=fwrisk!8},
    abox/.style={base, text width=34mm, rounded corners=5pt,
            draw=fwagent!70, fill=fwagent!8},
    pflow/.style={->, thick, draw=fwplanner!80},
    rflow/.style={->, thick, draw=fwrisk!85},
    aflow/.style={->, thick, draw=fwagent!80},
    query/.style={->, thick, draw=fwagent!75},
    lbl/.style={font=\scriptsize, align=center, inner sep=1.2pt, text=black!70},
    glbl/.style={font=\scriptsize\itshape, inner sep=1pt}
    ]

    % ---------------- group panels ----------------
    \draw[draw=fwplanner!45, dashed, rounded corners=3pt, fill=fwplanner!3]
    (-2.50,-0.45) rectangle (2.50,5.45);
    \node[glbl, text=fwplanner, anchor=north east] at (2.40,5.35) {Planner};

    \draw[draw=fwrisk!45, dashed, rounded corners=3pt, fill=fwrisk!3]
    (3.50,-0.45) rectangle (8.65,5.65);
    \node[glbl, text=fwrisk, anchor=north west] at (3.60,5.55) {Risk model};

    % ---------------- planner column (x = 0, east edge at x = 2.00) ----------------
    \node[abox] (snap)   at (0, 6.20) {Snapshot of the\\environment};
    \node[pbox] (sample) at (0, 4.50) {Sample and steer};
    \node[pbox] (edge)   at (0, 3.20) {Candidate edge};
    \node[pbox] (test)   at (0, 1.90) {Feasibility test};
    \node[pbox] (rewire) at (0, 0.30) {Rewire};
    \node[abox] (path)   at (0,-1.90) {Path passed to\\the agent};

    % ---------------- risk model column (x = 6.10, west edge at x = 3.80) --------
    \node[rbox] (ev)  at (6.10,4.40) {Evidence at the sample\\points along the edge};
    \node[rbox] (bbn) at (6.10,2.55) {Bayesian belief network};
    \node[rbox] (red) at (6.10,0.70) {Severity distributions\\aggregated to a scalar};

    % ---------------- flow inside the planner ----------------
    \draw[aflow] (snap)   -- (sample);
    \draw[pflow] (sample) -- (edge);
    \draw[pflow] (test)   -- (rewire);
    \node[lbl, anchor=west] at (0.15,1.10) {feasible};
    \draw[pflow] (rewire) -- (path);
    \node[lbl, anchor=east] at (-0.18,-0.85) {feasible and\\cost-improving};

    % ---------------- query out: run at y = 3.20, turn at x = 3.30 ----------
    \draw[query] (2.00,3.20) -- (3.30,3.20) -- (3.30,4.40) -- (3.80,4.40);

    \draw[rflow] (ev)  -- (bbn);
    \draw[rflow] (bbn) -- (red);

    % ---------------- query back: turn at x = 3.30, run at y = 1.90 ---------
    \draw[query] (3.80,0.70) -- (3.30,0.70) -- (3.30,1.90) -- (2.00,1.90);

    % ---------------- replanning loop ----------------
    \draw[aflow] (-2.00,-1.90) -- (-3.0,-1.90) -- (-3.0,6.00) -- (-2.00,6.00);
    \node[lbl, anchor=east] at (-3.30,2.20) {replan};

\end{tikzpicture}

%% file: diagrams/generic_severity_structure.tex
{
\tikzset{
  riskfig/.style={
      >=Stealth,
      font=\footnotesize,
      every node/.style={align=center, inner sep=2.5pt},
      haz/.style={draw, rounded corners=2pt, fill=red!22, minimum height=1.1cm, minimum width=3.4cm},
      sev/.style={draw, rounded corners=2pt, fill=yellow!45, minimum height=0.85cm, minimum width=3.4cm},
      box/.style={draw, rounded corners=3pt, fill=gray!8, inner sep=5pt},
      arr/.style={->, semithick, black!65},
      plate/.style={draw=black!55, dashed, rounded corners=3pt, inner sep=6pt},
      ghost/.style={draw=black!22, dashed, rounded corners=3pt, fill=white},
      note/.style={font=\scriptsize, text=black!55},
    }
}

\begin{tikzpicture}[riskfig]

  % Event risk index, one per hazardous event
  \node[haz] (rh) at (0, 0)
  {Event risk index\\[2pt] $r_h=\sum_{s} p_{h,s}\,\tilde{w}_h(s)$};

  % Plate with stacked copies
  \node[plate, fit=(rh)] (plate) {};
  \begin{scope}[on background layer]
    \draw[ghost] ($(plate.south west)+(0.40,0.40)$) rectangle ($(plate.north east)+(0.40,0.40)$);
    \draw[ghost] ($(plate.south west)+(0.20,0.20)$) rectangle ($(plate.north east)+(0.20,0.20)$);
    \fill[white, rounded corners=3pt] (plate.south west) rectangle (plate.north east);
  \end{scope}
  \node[note, anchor=north] at (plate.south) {one per hazardous event $h$};

  % Input from the BBN
  \draw[arr] ($(plate.west)+(-1.4,0)$) -- node[midway, above, font=\scriptsize] {$p_{h,s}$} (plate.west);

  % Combined risk index
  \node[box, anchor=west] (r) at ($(plate.east)+(1.6,0)$)
  {\textbf{Risk index}\\[3pt] $r=1-\prod_{h}(1-r_h)\in[0,1]$};
  \draw[arr] (plate.east) -- node[midway, above, font=\scriptsize] {$r_h$} (r.west);

  % Output to the planner
  \draw[arr] (r.east) -- node[midway, above, font=\scriptsize] {$r$} ++(1.4,0);

\end{tikzpicture}
}

%% file: diagrams/bbn_without_evidence.tex
% Requires: \usepackage{tikz}
\usetikzlibrary{arrows.meta, fit, backgrounds}

% Layout: collision chain up the left spine, grounding chain up the right.
% Each node shows its name on top and, below a rule, its states (\st).
% Ranked nodes also show their variance sigma^2 (\va).
% Dashed nodes are expanded in the companion evidence figures.
% Arc labels are the ranked-node weights of the child node (BBN_autonomous_ranked.xdsl).
\begin{tikzpicture}[
        >=Stealth,
        every node/.style={
                font=\scriptsize,
                align=center,
                inner sep=2pt,
                execute at begin node={\hyphenpenalty=10000 \exhyphenpenalty=10000},
            },
        common/.style={
                rectangle, draw,
                text width=2.9cm,
            },
        leaf/.style={common, fill=cyan!30},                        % root / evidence
        summary/.style={common, fill=cyan!15, dashed, thick},      % expanded elsewhere
        intermediate/.style={common, fill=white},                  % latent
        event/.style={common, fill=magenta!20},                    % accident
        outcome/.style={common, fill=orange!25, text width=3.4cm},                   % severity
        arr/.style={->, thick, color=blue!70},
        w/.style={fill=white, font=\tiny, inner sep=0.8pt, text=blue!50!black},  % ranked-node weight
        group/.style={draw=gray!55, dashed, rounded corners, inner sep=5pt},
        grouplabel/.style={font=\scriptsize\itshape, color=gray!70!black},
    ]

    % node states, written below a thin rule
    \def\st#1{\par\vspace{1pt}\hrule\vspace{1.5pt}{\tiny\itshape #1}}
    % ranked-node variance, written below a gray rule
    \def\va#1{\par\vspace{1pt}{\color{gray!60}\hrule}\vspace{1.5pt}{\tiny \mbox{$\sigma^2 = #1$}}}

    % ============================================================
    % Left spine: collision
    % ============================================================
    \node[intermediate] (failcoll) at (0, 0)
    {Failure of recovery from collision course\st{Likely, Unlikely}\va{0.032}};
    \node[event] (coll) at (0, 3.2)
    {Collision\st{Likely, Unlikely}\va{0.016}};
    \node[outcome] (collsev) at (0, 6.4)
    {Collision Risk\st{Negligible, Marginal, Moderate, Critical, Catastrophic}\va{0.091}};

    \draw[arr] (failcoll) -- node[w] {$10$} (coll);
    \draw[arr] (coll)     -- node[w] {$10$} (collsev);

    % --- collision-only parents, outboard of the spine ---
    \node[summary] (encprop) at (-3.6, 1.6)
    {Encounter properties\st{Negligible, Critical}\va{0.019}};
    \node[leaf] (targetves) at (-3.6, 4.8)
    {Target vessel fails to recover\st{Likely, Unlikely}};

    \draw[arr] (encprop)   -- node[w] {$1.5$} (failcoll);
    \draw[arr] (encprop)   -- node[w] {$10$} (coll);
    \draw[arr] (encprop)   -- node[w, pos=0.3] {$7$} (collsev);
    \draw[arr] (targetves) -- node[w] {$2.6$} (coll);

    % ============================================================
    % Right spine: grounding
    % ============================================================
    \node[intermediate] (failgr) at (8.0, 0)
    {Failure of recovery from grounding course\st{Yes, No}\va{0.034}};
    \node[event] (gr) at (8.0, 3.2)
    {Grounding\st{Likely, Unlikely}\va{0.01}};
    \node[outcome] (grsev) at (8.0, 6.4)
    {Grounding Risk\st{Negligible, Marginal, Moderate, Critical, Catastrophic}\va{0.091}};

    \draw[arr] (failgr) -- node[w] {$1$} (gr);
    \draw[arr] (gr)     -- node[w] {$6$} (grsev);

    % --- grounding-only parents, outboard of the spine ---
    \node[summary] (waterway) at (11.6, 1.6)
    {Waterway properties\st{Negligible, Critical}\va{0.046}};
    \node[leaf] (grcourse) at (11.6, 4.8)
    {Vessel on grounding course\st{Yes, No}};

    \draw[arr] (waterway) -- node[w, pos=0.65] {$5.5$} (failgr);
    \draw[arr] (waterway) -- node[w, pos=0.3] {$6$} (grsev);
    \draw[arr] (waterway) -- node[w, pos=0.7] {$1$} (failcoll);
    \draw[arr] (grcourse) -- node[w, pos=0.22] {$1$} (gr);

    % ============================================================
    % Central column: parents shared by both branches
    % ============================================================
    \node[leaf] (tech) at (2.2, -3.9)
    {Technical failure\st{Likely, Unlikely}};
    \node[leaf] (human) at (5.8, -3.9)
    {Human error\st{Likely, Unlikely}};
    \node[intermediate] (ownves) at (4.0, -1.8)
    {Own vessel fails to recover\st{Likely, Unlikely}\va{0.0052}};
    \node[summary] (envprop) at (4.0, 3.2)
    {Environment properties\st{Negligible, Critical}\va{0.077}};
    \node[leaf] (rescue) at (4.0, 7.6)
    {Available rescue resources\st{Yes, No}};

    \draw[arr] (tech)   -- node[w] {$2.8$} (ownves);
    \draw[arr] (human)  -- node[w] {$1.85$} (ownves);
    \draw[arr] (ownves) -- node[w] {$3.45$} (failcoll);
    \draw[arr] (ownves) -- node[w] {$4.25$} (failgr);

    \draw[arr] (envprop) -- node[w, pos=0.4] {$1$} (failcoll);
    \draw[arr] (envprop) -- node[w, pos=0.4] {$0.05$} (failgr);
    \draw[arr] (envprop) -- node[w, pos=0.4] {$4$} (collsev);
    \draw[arr] (envprop) -- node[w, pos=0.4] {$3$} (grsev);

    \draw[arr] (rescue) -- node[w] {$1$} (collsev);
    \draw[arr] (rescue) -- node[w] {$1$} (grsev);

    \begin{scope}[on background layer]
        \node[group, fit=(encprop)] (boxA) {};
        \node[group, fit=(envprop)] (boxB) {};
        \node[group, fit=(waterway)] (boxC) {};
    \end{scope}

    \node[grouplabel, anchor=south west] at (boxA.north west) {(a)};
    \node[grouplabel, anchor=south west] at (boxB.north west) {(b)};
    \node[grouplabel, anchor=south west] at (boxC.north west) {(c)};

\end{tikzpicture}

%% file: diagrams/bbn_encounter_properties.tex
% Requires: \usepackage{tikz}
\usetikzlibrary{arrows.meta, fit, backgrounds}

% Layout: parents fan out in an arc below their child ("flower").
% Each node shows its name on top and, below a rule, its states (\st).
% Ranked nodes also show their variance sigma^2 (\va).
% Numbered evidence states (State0 ... State4) are written as evidence_{0,...,4}.
% Arc labels are the ranked-node weights of the child node (BBN_autonomous_ranked.xdsl).
\begin{tikzpicture}[
        >=Stealth,
        every node/.style={
                font=\scriptsize,
                align=center,
                inner sep=2pt,
                execute at begin node={\hyphenpenalty=10000 \exhyphenpenalty=10000},
            },
        common/.style={
                rectangle, draw,
                text width=2.1cm,
            },
        intermediate/.style={common, fill=white},
        leaf/.style={common, fill=cyan!30},
        arr/.style={->, thick, color=blue!70},
        w/.style={fill=white, font=\tiny, inner sep=0.8pt, text=blue!50!black},  % ranked-node weight
        group/.style={draw=gray!55, dashed, rounded corners, inner sep=5pt},
        grouplabel/.style={font=\scriptsize\itshape, color=gray!70!black},
    ]

    % node states, written below a thin rule
    \def\st#1{\par\vspace{1pt}\hrule\vspace{1.5pt}{\tiny\itshape #1}}
    % ranked-node variance, written below a gray rule
    \def\va#1{\par\vspace{1pt}{\color{gray!60}\hrule}\vspace{1.5pt}{\tiny \mbox{$\sigma^2 = #1$}}}

    % ============================================================
    % (a) Encounter properties  -- 4 petals
    % ============================================================
    \node[intermediate, fill=cyan!15, dashed, thick] (encprop) at (0, 0)
    {Encounter properties\st{Negligible, Critical}\va{0.019}};

    \node[leaf] (relvel) at (-3.6, -1.5) {Relative velocity\st{$\textit{evidence}_{0, \cdots, 4}$}};
    \node[leaf] (tcpa)   at (-1.2, -2.9) {TCPA\st{$\textit{evidence}_{0, \cdots, 4}$}};
    \node[leaf] (dcpa)   at ( 1.2, -2.9) {DCPA\st{$\textit{evidence}_{0, \cdots, 4}$}};
    \node[leaf] (enc)    at ( 3.6, -1.5) {Encounter\st{Likely, Unlikely}};

    \draw[arr] (relvel) -- node[w] {$1.35$} (encprop);
    \draw[arr] (tcpa)   -- node[w] {$1.75$} (encprop);
    \draw[arr] (dcpa)   -- node[w] {$2.2$} (encprop);
    \draw[arr] (enc)    -- node[w] {$10$} (encprop);

    \begin{scope}[on background layer]
        \node[group, fit=(encprop)(relvel)(tcpa)(dcpa)(enc)] (boxA) {};
    \end{scope}

    \node[grouplabel, anchor=north west] at (boxA.north west) {(a)};

\end{tikzpicture}

%% file: diagrams/bbn_environment_properties.tex
% Requires: \usepackage{tikz}
\usetikzlibrary{arrows.meta, fit, backgrounds}

% Layout: parents fan out in an arc below their child ("flower").
% Each node shows its name on top and, below a rule, its states (\st).
% Ranked nodes also show their variance sigma^2 (\va).
% Numbered evidence states (State0 ... State4) are written as evidence_{0,...,4}.
% Arc labels are the ranked-node weights of the child node (BBN_autonomous_ranked.xdsl).
\begin{tikzpicture}[
        >=Stealth,
        every node/.style={
                font=\scriptsize,
                align=center,
                inner sep=2pt,
                execute at begin node={\hyphenpenalty=10000 \exhyphenpenalty=10000},
            },
        common/.style={
                rectangle, draw,
                text width=2.1cm,
            },
        intermediate/.style={common, fill=white},
        leaf/.style={common, fill=cyan!30},
        arr/.style={->, thick, color=blue!70},
        w/.style={fill=white, font=\tiny, inner sep=0.8pt, text=blue!50!black},  % ranked-node weight
        group/.style={draw=gray!55, dashed, rounded corners, inner sep=5pt},
        grouplabel/.style={font=\scriptsize\itshape, color=gray!70!black},
    ]

    % node states, written below a thin rule
    \def\st#1{\par\vspace{1pt}\hrule\vspace{1.5pt}{\tiny\itshape #1}}
    % ranked-node variance, written below a gray rule
    \def\va#1{\par\vspace{1pt}{\color{gray!60}\hrule}\vspace{1.5pt}{\tiny \mbox{$\sigma^2 = #1$}}}

    % ============================================================
    % (b) Environment properties  -- 3 petals
    % ============================================================
    \node[intermediate, fill=cyan!15, dashed, thick] (envprop) at (0, 0)
        {Environment properties\st{Negligible, Critical}\va{0.077}};

    \node[leaf] (wind) at (-2.7, -1.6) {Wind speed\st{$\textit{evidence}_{0, \cdots, 4}$}};
    \node[leaf] (sea)  at ( 0.0, -3.0) {Sea state\st{$\textit{evidence}_{0, \cdots, 4}$}};
    \node[leaf] (vis)  at ( 2.7, -1.6) {Visibility\st{$\textit{evidence}_{0, \cdots, 4}$}};

    \draw[arr] (wind) -- node[w] {$1.15$} (envprop);
    \draw[arr] (sea)  -- node[w] {$1$} (envprop);
    \draw[arr] (vis)  -- node[w] {$1.15$} (envprop);

    \begin{scope}[on background layer]
        \node[group, fit=(envprop)(wind)(sea)(vis)] (boxB) {};
    \end{scope}

    \node[grouplabel, anchor=north west] at (boxB.north west) {(b)};

\end{tikzpicture}

%% file: diagrams/bbn_waterway_properties.tex
% Requires: \usepackage{tikz}
\usetikzlibrary{arrows.meta, fit, backgrounds}

% Layout: parents fan out in an arc below their child ("flower").
% Each node shows its name on top and, below a rule, its states (\st).
% Ranked nodes also show their variance sigma^2 (\va).
% Numbered evidence states (State0 ... State4) are written as evidence_{0,...,4}.
% Arc labels are the ranked-node weights of the child node (BBN_autonomous_ranked.xdsl).
\begin{tikzpicture}[
        >=Stealth,
        every node/.style={
                font=\scriptsize,
                align=center,
                inner sep=2pt,
                execute at begin node={\hyphenpenalty=10000 \exhyphenpenalty=10000},
            },
        common/.style={
                rectangle, draw,
                text width=2.1cm,
            },
        intermediate/.style={common, fill=white},
        leaf/.style={common, fill=cyan!30},
        arr/.style={->, thick, color=blue!70},
        w/.style={fill=white, font=\tiny, inner sep=0.8pt, text=blue!50!black},  % ranked-node weight
        group/.style={draw=gray!55, dashed, rounded corners, inner sep=5pt},
        grouplabel/.style={font=\scriptsize\itshape, color=gray!70!black},
    ]

    % node states, written below a thin rule
    \def\st#1{\par\vspace{1pt}\hrule\vspace{1.5pt}{\tiny\itshape #1}}
    % ranked-node variance, written below a gray rule
    \def\va#1{\par\vspace{1pt}{\color{gray!60}\hrule}\vspace{1.5pt}{\tiny \mbox{$\sigma^2 = #1$}}}

    % ============================================================
    % (c) Waterway properties  -- 3 petals
    % ============================================================
    \node[intermediate, fill=cyan!15, dashed, thick] (watprop) at (0, 0)
    {Waterway properties\st{Negligible, Critical}\va{0.046}};

    \node[leaf] (dist)    at (-2.7, -1.6) {Distance to shore\st{$\textit{evidence}_{0, \cdots, 4}$}};
    \node[leaf] (depth)   at ( 0.0, -3.0) {Depth\st{$\textit{evidence}_{0, \cdots, 4}$}};
    \node[leaf] (traffic) at ( 2.7, -1.6) {Ship traffic\st{Negligible, Critical}};

    \draw[arr] (dist)    -- node[w] {$7$} (watprop);
    \draw[arr] (depth)   -- node[w] {$3.5$} (watprop);
    \draw[arr] (traffic) -- node[w] {$2.45$} (watprop);

    \begin{scope}[on background layer]
        \node[group, fit=(watprop)(dist)(depth)(traffic)] (boxC) {};
    \end{scope}

    \node[grouplabel, anchor=north west] at (boxC.north west) {(c)};

\end{tikzpicture}